\documentclass[sn-apa, iicol]{sn-jnl}

\usepackage{graphicx}%
\usepackage{multirow}%
\usepackage{amsmath,amssymb,amsfonts}%
\usepackage{amsthm}%
\usepackage{mathrsfs}%
\usepackage[title]{appendix}%
\usepackage{xcolor}%
\usepackage{textcomp}%
\usepackage{manyfoot}%
\usepackage{booktabs}%
\usepackage{algorithm}%
\usepackage{algorithmicx}%
\usepackage{algpseudocode}%
\usepackage{listings}%
\usepackage{amssymb} 
\usepackage{pifont}
\definecolor{mssdgred}{rgb}{1.0, 0.2, 0.2}
\definecolor{ttpablue}{rgb}{0.2, 0.6, 1.0}
\definecolor{alwaysgreen}{rgb}{0.2, 0.8, 0.4}

\theoremstyle{thmstyleone}%
\theoremstyle{thmstyletwo}%

\theoremstyle{thmstylethree}%

\begin{document}

\title[Align the GAP: Prior-based Unified Multi-Task Remote Physiological Measurement Framework For Domain Generalization and Personalization]{Align the GAP: Prior-based Unified Multi-Task Remote Physiological Measurement Framework For Domain Generalization and Personalization}


\author[1]{\fnm{Jiyao} \sur{Wang}}\email{jwanggo@connect.ust.hk}

\author[1]{\fnm{Xiao} \sur{Yang}}\email{xyang856@connect.hkust-gz.edu.cn}

\author[2]{\fnm{Hao} \sur{Lu}}\email{hlu585@connect.ust.hk}

\author*[1]{\fnm{Dengbo} \sur{He}}\email{dengbohe@hkust-gz.edu.cn}

\author[2]{\fnm{Kaishun} \sur{Wu}}\email{wuks@hkust-gz.edu.cn}

\affil[1]{\orgdiv{Systems Hub}, \orgname{The Hong Kong University of Science and Technology (Guangzhou)}, \orgaddress{\city{Guangzhou}, \state{Guangdong}, \country{China}}}

\affil[2]{\orgdiv{Information Hub}, \orgname{The Hong Kong University of Science and Technology (Guangzhou)}, \orgaddress{\city{Guangzhou}, \state{Guangdong}, \country{China}}}


\abstract{

Multi-source synsemantic domain generalization (MSSDG) for multi-task remote physiological measurement seeks to enhance the generalizability of these metrics and attracts increasing attention. However, challenges like partial labeling and environmental noise may disrupt task-specific accuracy. Meanwhile, given that real-time adaptation is necessary for personalized products, the test-time personalized adaptation (TTPA) after MSSDG is also worth exploring, while the gap between previous generalization and personalization methods is significant and hard to fuse. Thus, we proposed a unified framework for MSSD\textbf{G} and TTP\textbf{A} employing \textbf{P}riors (\textbf{GAP}) in biometrics and remote photoplethysmography (rPPG). We first disentangled information from face videos into invariant semantics, individual bias, and noise. Then, multiple modules incorporating priors and our observations were applied in different stages and for different facial information. Then, based on the different principles of achieving generalization and personalization, our framework could simultaneously address MSSDG and TTPA under multi-task remote physiological estimation with minimal adjustments. We expanded the MSSDG benchmark to the TTPA protocol on six publicly available datasets and introduced a new real-world driving dataset with complete labeling. Extensive experiments that validated our approach, and the codes along with the new dataset will be released.
}




\maketitle

\section{Introduction}\label{sec1}

Vital signs like heart rate (HR), respiratory rate (RR), and blood oxygen saturation     (SpO2) are crucial indicators of health and sympathetic activation. These metrics are vital for various applications, including affective computing \citep{lu2024gpt} and driver state monitoring \citep{wang2024efficient, wang2024association}. Traditional methods mainly use contact-based sensors like Electrocardiography (ECG) and Photoplethysmography (PPG), which can hardly be used in contact-free settings. Recently, remote photoplethysmography (rPPG) has emerged, which can use low-cost webcams to measure cardiovascular health \citep{verkruysse2008remote,braun2024suboptimal} by detecting light absorption changes in facial skin. It can also estimate RR from movement and skin light reflection and SpO2 based on channel-wise light absorption differences.

\begin{figure*}
\begin{center}
\includegraphics[scale=0.45]{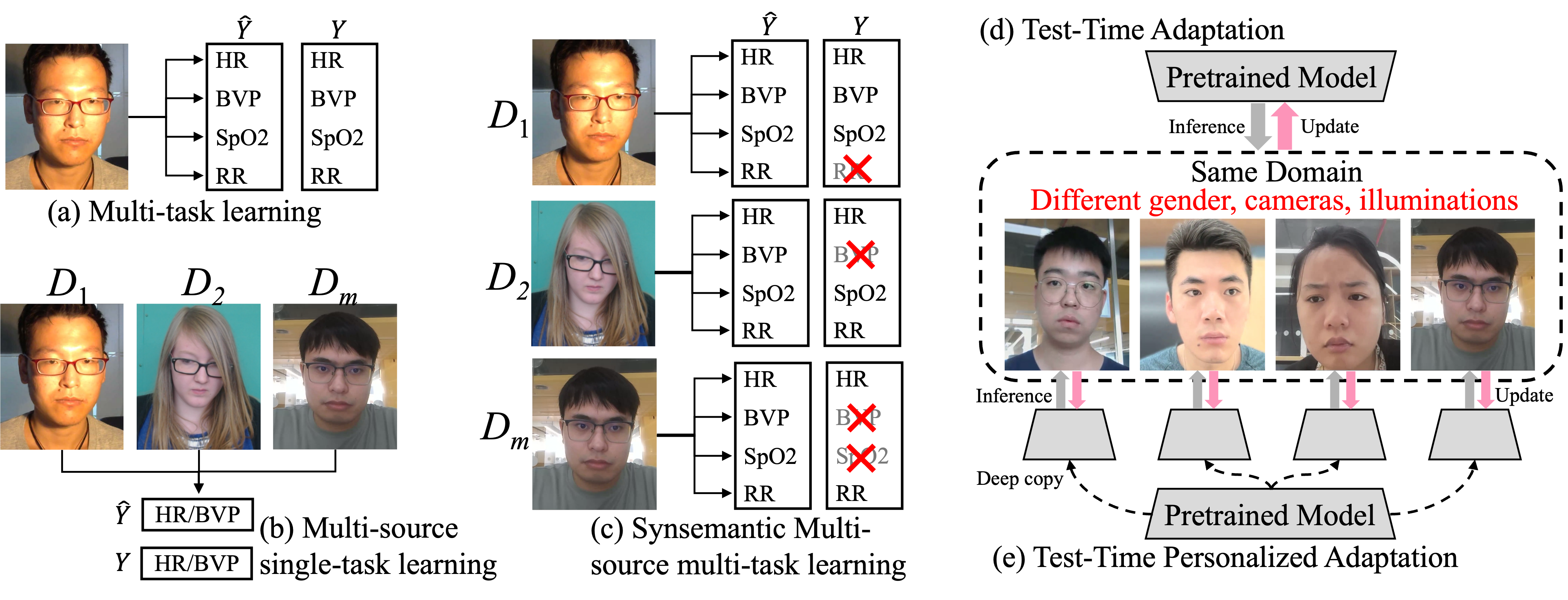}\\
\end{center}
\caption{Illustration of the difference between (a) classic multi-task learning. It assumes a single data source and jointly learns multiple tasks; (b) multi-source single-task learning, which aggregates data from multiple domains but only optimizes for a single task; (c) synsemantic multi-source multi-task learning. It simultaneously leverages multiple tasks and multiple domains, but needs to face imbalanced label spaces (d) test-time adaptation, which unsupervisely adapts a pretrained model to a new domain during inference, but it typically treats all test samples uniformly without considering individual differences; (e) test-time personalized adaptation, which further extends test-time adaptation by tailoring the model to different individuals who might be in the same environment. }\label{f1}
\end{figure*}

In recent years, remote physiological measurement methodology has advanced from handcrafted techniques \citep{de2013robust, tarassenko2014non} to deep learning (DL) methods \citep{yu2019remote, niu2019rhythmnet, yu2023physformer++, narayanswamy2024bigsmall}. Conventional approaches require assumptions or manual signal extraction for different tasks \citep{wang2016algorithmic}. However, relying on a single physiological indicator is often insufficient for comprehensive medical diagnosis \citep{Orphanidou2015SignalQualityIF}. Using multiple single-task models raises deployment costs and overlooks interdependencies among physiological indicators \citep{chang2023pepnet}. For example, respiratory variations can influence blood flow, affecting the peripheral vasculature and blood volume pulse (BVP) signal \citep{peper2007there}. In contrast, multi-task learning (MTL) approaches \citep{liu2020multi, narayanswamy2024bigsmall} automate facial feature extraction with a unified model, reducing reliance on independent assumptions and leveraging complementary features to enhance performance.

\begin{table*}[h]
\label{t1}
\centering
\caption{Comparison of Different Protocols. In this table,
$X^s$ and $Y^s$ mean the input and label space of the source domain, and $X^t, Y^t$ are for the target domain. The $X^s$ in MSDG and MSSDG consist of multiple domains of $X^s_i$, and there are multiple partially labeled output spaces $Y_i^{s*}$ in MSSDG. In TTPA, $X^t_i$ indicates that only one-subject data is available for each time of adaptation.}
\scriptsize
\begin{tabular}{lccccccl}
\toprule
\multirow{2}{*}{Protocol} & \multicolumn{4}{c}{Training} & \multicolumn{1}{c}{\multirow{2}{*}{Backward phase}}&\multicolumn{1}{c}{\multirow{2}{*}{Evaluation}} \\
\cmidrule(lr){2-5}
 & Source data & Source label & Target data & Target label & & \\
\midrule
DA & $X^s$ & $Y^s$ & $X^t$ & $Y^t$ & Training & Domain-specific \\
UDA & $X^s$ & $Y^s$ & $X^t$ & \_& Training & Domain-specific \\
SFDA & \_ & \_ & $X^t$ & \_ & Training& Domain-specific \\
DG & $X^s$ & $Y^s$ & \_ & \_ & Training& Domain-specific \\
MSDG & $X^s=\{X_i^s\}_{i=1}^D$ & $Y^s$ & \_ & \_ & Training& Domain-specific \\
MSSDG &  $X^s=\{X_i^s\}_{i=1}^D$ & $Y^{s*}=\{Y_i^{s*}\}_{i=1}^D$ & \_ & \_ & Training& Domain-specific \\
TTA & \_ & \_ & $X^t$ & \_ & Testing& Domain-specific \\
TTPA & \_ & \_ & $X^t_i$ & \_ & Testing& Person-specific \\
\bottomrule
\end{tabular}
\end{table*}

Despite advancements, previous MTL methods trained on multi-source datasets struggle with generalization in unseen domains \citep{wang2024physmle}. Domain shifts in rPPG stem from significant variations among individuals or environmental conditions, which are always regarded as noise that obscures semantic information. These discrepancies can persist even in controlled settings, and can become worse in unpredictable real-world environments. Recently, there has been growing attention on improving multi-source domain generalization (MSDG) in rPPG \citep{lu2023neuron, wang2023hierarchical, 10.1145/3581783.3612265, wang2024generalizable}. However, beyond typical MSDG challenges, unique difficulties arise from the interactions between physiological signals. In other words, current MSDG methods mostly target single-task estimation (e.g., HR or BVP) and have not been adapted to Multi-source Synsemantic Domain Generalization (MSSDG) \citep{wang2024physmle}, where, in addition to the need to address MTL with multi-domain shift, there is also a need to overcome the severe seesaw effect of partial labels across source domains (Figure \ref{f1}(a) to (c)). 

To excel in deployment environments, post-DG with effective adaptation in the training process is essential. As shown in Figure \ref{f1}(d), Test-time adaptation (TTA) adjusts the domain-agnostic optimal point set by DG methods to the optimal point of the target domain with unlabeled data \citep{10552776}. Unfortunately, current TTA methods \citep{li2024bi,huang2024fully} for rPPG overlook individual characteristics and environment complexities and have not been tailored for MTL settings. Thus, instead of solely focusing on adapting to the target domain, test-time personalized adaptation (TTPA) is crucial for rPPG (Figure \ref{f1}(e)). Additionally, due to the varied goals of generalization and personalization, previous MSSDG and TTA methods cannot align and need separate designs, increasing the efforts during training.

To tackle these issues, this paper aims to achieve a unified framework for MSSD\textbf{G} and TTP\textbf{A} with \textbf{P}riors in biometrics and rPPG for remote multi-task physiological measurement (\textbf{GAP}). In general, based on our observations and domain-related knowledge, we noticed the consistency differences between physiological signals and individual differences in visual cues and vital signs in the time and frequency domains. Thus, we designed a series of shared approaches to resolve challenges in both MSSDG and TTPA. First, the partial annotation of certain tasks can lead to insufficient learning and the seesaw effect, which hinders generalization during supervised training in MSSDG. Instead of relying on feature-level augmentation \citep{wang2024physmle} or mixup \citep{zhou2024mixstyle}, we introduced a novel input-level data augmentation method based on the time-domain inconsistency and the principle of Ratio-of-Ratios (RoR) method of SpO2 estimation \citep{berntson1997heart,tarassenko2014non}. We adjusted the pulsatile absorbance ratios of different color channels in facial videos to simulate environmental noises where the same individual remains in the same physiological state.

Then, a novel self-supervised regularization system was constructed based on the properties of biometrics and human facial cues. To enhance shared representation learning stability amidst semantic-irrelevant noise, we proposed two novel self-supervised regularizations focusing on the semantic structure and distribution of low-level features from a shared feature encoder, respectively. Additionally, to address the lack of reliable labels in both MSSDG and TTPA, we introduced a self-supervised loss combination with the spatio-temporal frequency domain consistency and time domain inconsistency of physiological measurements to narrow the effective learning range for generalization and personalization. Further, based on the individual heterogeneity we identified, we explicitly modeled individual bias through an auxiliary task branch. Through the design of the individual difference disentangling and fusion module, the auxiliary task and the intermediate features it forms (which represent individual information) can be flexibly used for individual bias elimination or personalization self-adaptation in the MTL setting. We summarize the contribution of this work as follows:

\begin{itemize}
    \item We proposed a novel unified framework GAP for multi-task remote physiological measurement, which was designed based on multiple expert knowledge and experimental observations to achieve generalization and personalization.
    \item To address the prevalent problem of insufficient labels and the effect of environmental noises in both MSSDG and TTPA, we introduced several self-supervised losses based on the spatio-temporal frequency-domain consistency of physiological signals and semantic structure and distribution robust alignment, respectively.
    \item We experimentally validated the impact of individual differences on physiological monitoring and introduced a flexible and effective individual difference explicit modeling scheme that enables individual difference elimination or enhancement depending on the fitted target. As far as we know, it is the first to instantiate the time-domain inconsistency in the rPPG task for individual bias learning. 
    \item As the follow-up study of \citep{wang2024physmle} and the first study to achieve personalization in multi-task remote physiological measurement, we validated the effectiveness of the proposed GAP over six datasets, which further extended the previous MSSDG benchmark with TTPA parts. Besides, to further verify the performance in the real-world application, we collect and publicize a new dataset (HMPC-D). Extensive experiments showed that our method worked well in MSSDG and TTPA protocols.
\end{itemize}




\section{Related Work}\label{sec2}
\subsection{Remote Physiological Measurement}

Since 2008, rPPG has become a key non-contact method for physiological measurement \citep{verkruysse2008remote}. The green channel provides the strongest plethysmography signal due to the absorption of hemoglobin. This technique uses optical principles like Lambert-Beer law and Shafer’s dichromatic reflection model to analyze light interaction with skin \citep{tarassenko2014non}. Traditional rPPG methods depend on statistical analysis of heartbeat data and manual adjustments, which are complicated by variations in skin tones and lighting. Recent deep learning advancements have enhanced accuracy, as seen in the DeepPhys model \citep{chen2018deepphys}, which uses CNNs to extract BVP signals. Other approaches include 3D-CNN \citep{yu2019remote} and Transformers \citep{yu2023physformer++}.

Besides HR, RR is crucial for health monitoring. Earlier studies assessed RR through pixel movements \citep{cheng2023motion} or intensity changes \citep{janssen2015video}. Both traditional rPPG \citep{ghodratigohar2019remote} and deep learning methods \citep{liu2020multi, narayanswamy2024bigsmall} have been applied for RR estimation, yet issues like motion artifacts and non-periodic respiration can lead to inaccuracies \citep{du2021weakly}. Traditionally, rPPG focused on single outputs like HR. However, newer models like MTTS-CAN and BigSmall allow real-time cardiovascular and respiratory measurements, also capturing facial movements and HR \citep{das2021bvpnet,du2021weakly}. SpO2 has usually been measured with specialized cameras for DC and AC components \citep{kong2013non} and recent studies show promise in estimating SpO2 using standard RGB cameras by observing skin color changes \citep{bal2015non,tarassenko2014non}. Emerging deep learning methods \citep{hu2023contactless,akamatsu2023blood} for contactless SpO2 estimation are promising, but achieving clinic-level accuracy remains challenging. Additionally, prior multi-task rPPG research was often trained on single domains, and publicly available datasets typically provide limited labels.

\subsection{Domain Generalization in Remote Photoplethysmography}

Variability across domains hinders model performance on specific datasets \citep{lu2023neuron}. To address this, various DG techniques have been applied to rPPG applications \citep{du2023dual,wang2024generalizable,wang2023hierarchical}. The main goal of these DG methods is for models trained on source domain data to generalize effectively to unseen target domains. Three main categories exist: Data manipulation, which enhances the dataset via augmentation and synthetic data, increasing its variability \citep{shi2020towards,zhou2024mixstyle}; Representation learning, which extracts domain-invariant features for improved adaptability \citep{ganin2015unsupervised,jiyao2023DGrppg}; and Meta-learning, which prepares models for domain shifts through variations in virtual data \citep{finn2017model,lv2022causality}.

In rPPG, NEST \citep{lu2023neuron} optimizes the intermediate feature space during training but overlooks instance-specific variations, leading to poor performance with unique samples. Conversely, the HSRD framework \citep{wang2023hierarchical} separates domain-invariant features from instance-specific traits in BVP signals but relies on explicit domain labels. Overall, most DG methods \citep{wang2024generalizable, 10.1145/3581783.3612265} in rPPG mainly focused on HR and BVP signals. Expanding DG to a multi-task framework faces challenges due to semantic discrepancies and imbalances and thus negatively affects performance. Recently, PhysMLE \citep{wang2024physmle} combined low-rank adapters and tailored priors for MSSDG but required many hyperparameter adjustments and thus it is not easy to be applied to TTA scenarios.

\subsection{Test-time Adaptation in Remote Photoplethysmography}

Domain adaptation (DA) \citep{wang2018deep} has consistently attracted attention in previous research, particularly unsupervised DA (UDA) \citep{ganin2015unsupervised} and source-free DA \citep{yang2021generalized}, due to challenges in obtaining labeled data in remote physiological measurement contexts. Treating the deployment environment as a whole neglects individual user differences and complicates real-world data capture. Therefore, TTA is appealing thanks to its continuous gradient updating in small batches \citep{10552776} as it can adjust pre-trained models to fit the target domain during testing using domain-specific inputs \citep{zhao2023delta,wang2020tent}. Current TTA techniques can be categorized into Batch Normalization (BN) calibration \citep{schneider2020improving,nado2020evaluating,duboudin2022learning,bahmani2022semantic,klingner2022continual}, meta-learning strategies \citep{min2023meta,liu2022towards,chi2021test,soh2020meta}, and self-supervised learning methods \citep{zhang2020inference,li2021test}. 

Recently, TTA has been applied in rPPG tasks, with notable implementations leveraging self-supervised consistency priors \citep{li2024bi} and synthetic signals for adaptation \citep{huang2024fully}. While DG focuses on invariant feature identification across domains \citep{wang2023hierarchical}, TTA aims at specific context adaptation. A focus on consistent information in TTA may hinder capturing scene-specific details. Previous TTA methods ignored user heterogeneity, adapting the entire domain, which is unsuitable for the downstream application of rPPG. Thus, we aim to achieve personalization (TTPA) by further optimizing the  TTA.

\section{Methodology}
In this paper, we present a unified framework for both MSSDG and TTPA protocols in MTL rPPG tasks and leverage expert knowledge (GAP). We begin by formulating our research question in Section \ref{subsection: Definitions}. Next, in Section \ref{subsection: Observations}, we discuss our observations regarding rPPG, which serve as the foundation for the three key prior knowledge elements used in our proposal. Following that, we introduce each specific design in our proposed GAP framework, which consists of prior-based augmentation, robust shared representation learning, and adaptive synsemantic learning. Specifically, we propose the prior-based augmentation in Section \ref{subsection: Augmentation}. Then, we introduce the regularization designs to promote the robustness of learned shared representations in Section \ref{subsection: Shared}, and the proposed method tailored to synsemantic learning in Section \ref{subsection: Synsemantic}. Finally, the optimization and inference processes of our framework in MSSDG and TTPA are introduced in Section \ref{subsection: Optimization}.

\subsection{Problem Definitions}
\label{subsection: Definitions}

\begin{figure}
\begin{center}
\includegraphics[scale=0.46]{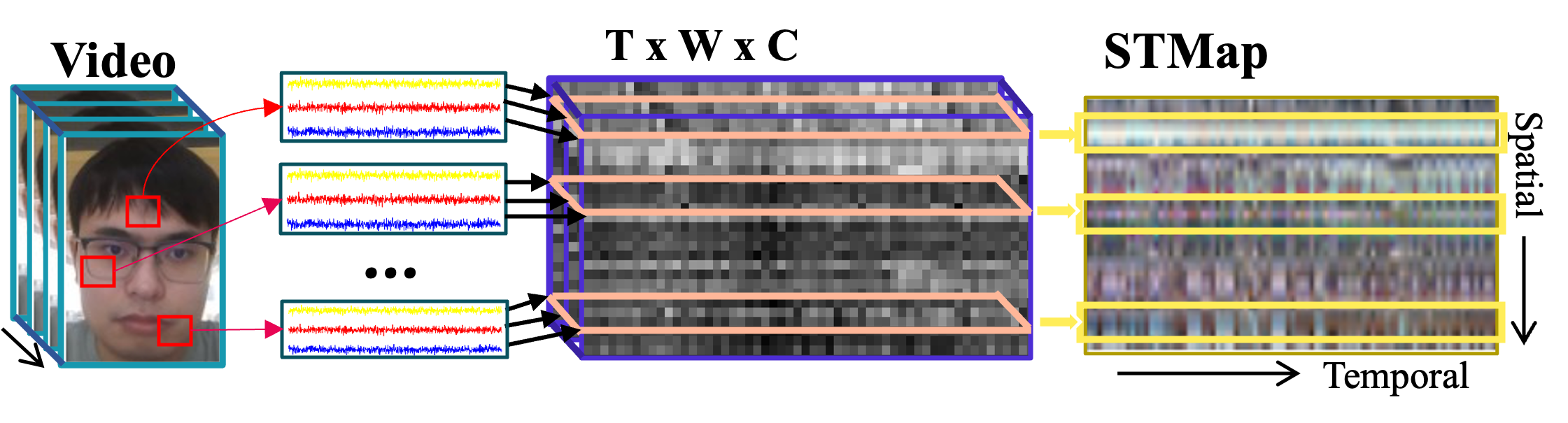}\\
\end{center}
\caption{The illustration of producing STMap from the facial RGB video.}\label{fstmap}
\end{figure}

We define the notations and problem settings for our study as follows. As illustrated in Figure \ref{f2}, we transformed a batch of $B$ raw facial RGB videos collected from multiple datasets into spatial-temporal maps (STMaps) \citep{niu2019rhythmnet} (Figure \ref{fstmap}). The resulting STMaps, denoted as $X = \{x_i\}{i=1}^{B}, x_i \in \mathbb{R}^{T \times W \times 3}$, were then fed into an MTL model, which is represented as $F(X; \theta): X\rightarrow Y$. Here, $T$ and $W$ refer to the length, and width of the STMaps, while $\theta$ represents the under-optimized parameters of the model. In this study, there are $N=4$ vital signs for estimation, denoted as $Y = \{y_{hr}, y_{bvp}, y_{spo}, y_{rr}\}$. Specifically, $y_{hr}$ represents HR, $y_{bvp}$ corresponds to BVP, and $y_{spo}$ and $y_{rr}$ indicate SpO2 and RR, respectively. According to the Bayes theorem and our preliminary work \citep{zhang2024advancing}, we can separate the $X$ into two independent components: semantic $X_{phys}$ (i.e., periodic facial color changes due to physiologic state) and noise $X_{noise}$ (e.g., illumination, skin color, camera, motion). Further, the model can also be formulated as:
\begin{align}
\mathcal{P}(Y|X) &= \frac{\mathcal{P}(X|Y)}{\mathcal{P}(X)} \cdot \mathcal{P}(Y) \nonumber \\
&= \frac{\mathcal{P}(X_{phy}, X_{noise}|Y)}{\mathcal{P}(X_{phy}, X_{noise})} \cdot \mathcal{P}(Y) \nonumber \\
&= \underbrace{\frac{\mathcal{P}(X_{phy}|Y)}{\mathcal{P}(X_{phy})}}_{\text{Semantic}} \cdot \underbrace{\frac{\mathcal{P}(X_{noise}|Y, X_{phy})}{\mathcal{P}(X_{noise}|X_{phy})}}_{\text{Prejudiced}} \cdot \mathcal{P}(Y) \label{eq1}
\end{align}



\subsection{Observations and Analysis}
\label{subsection: Observations}

\subsubsection{Individual Bias}

\begin{figure}
\begin{center}
\includegraphics[scale=0.1]{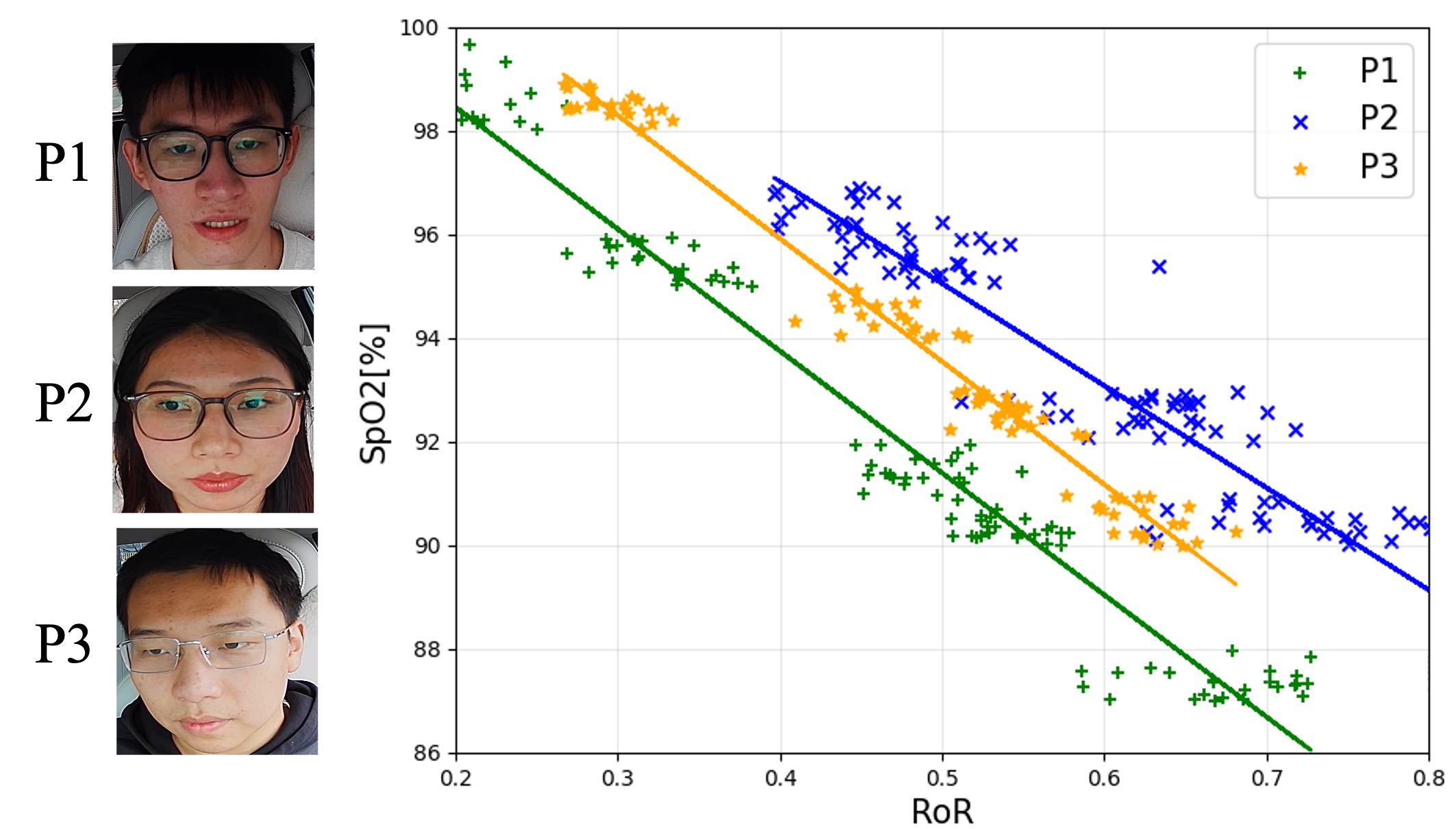}\\
\end{center}
\caption{The distribution between RoR and SpO2 level of three participants.}\label{f2}
\end{figure}

From Eq. (\ref{eq1}), for both generalizability and personalization, the goal was to strengthen the semantic part and weaken the effect of noise. In fact, in the semantic part, it is actually possible to further disentangle the semantic relationship between the truly invariant optical features and the physiological signals, as well as the amount of bias due to individual differences. For example, in oximetry, based on the Ratio-of-Ratios (RoR)-based method, we can calculate the absorbance difference of various color channels for each individual within a timeslot by using Eq. (\ref{eq2}). 

\begin{equation}
 R_\lambda = \frac{AC_\lambda}{DC_\lambda}, \quad RoR_{\lambda1,\lambda2} = \frac{AC_{\lambda1}/DC_{\lambda1}}{AC_{\lambda2}/DC_{\lambda2}} \label{eq2}
\end{equation}

The non-pulsating component (DC) is measured by the mean value, while the pulsating component (AC) is the standard deviation. $\lambda$ denotes the color channel. Previous studies \citep{guazzi2015non} indicate this ratio difference correlates linearly with SpO2 levels. Thus, we instructed 3 participants to rest for 1 minute and hold their breath for 30 seconds in the environment with constant lighting. Their faces were photographed with the ORBBEC Gemini Pro RGB camera at 30 fps, and SpO2 levels were recorded with WS20A\footnote{https://www.accbiomed.com/eproducts/list113.html}. After calculating the RoR per second as described in \citep{guazzi2015non}, we fitted the corresponding SpO2 values for each participant using one-way linear regression, shown in Figure \ref{f2}. Significant variations in intercepts among participants suggest individuals exhibit different RoRs for identical SpO2 levels, emphasizing the importance of user variability alongside domain differences.

\begin{figure}
\begin{center}
\includegraphics[scale=0.37]{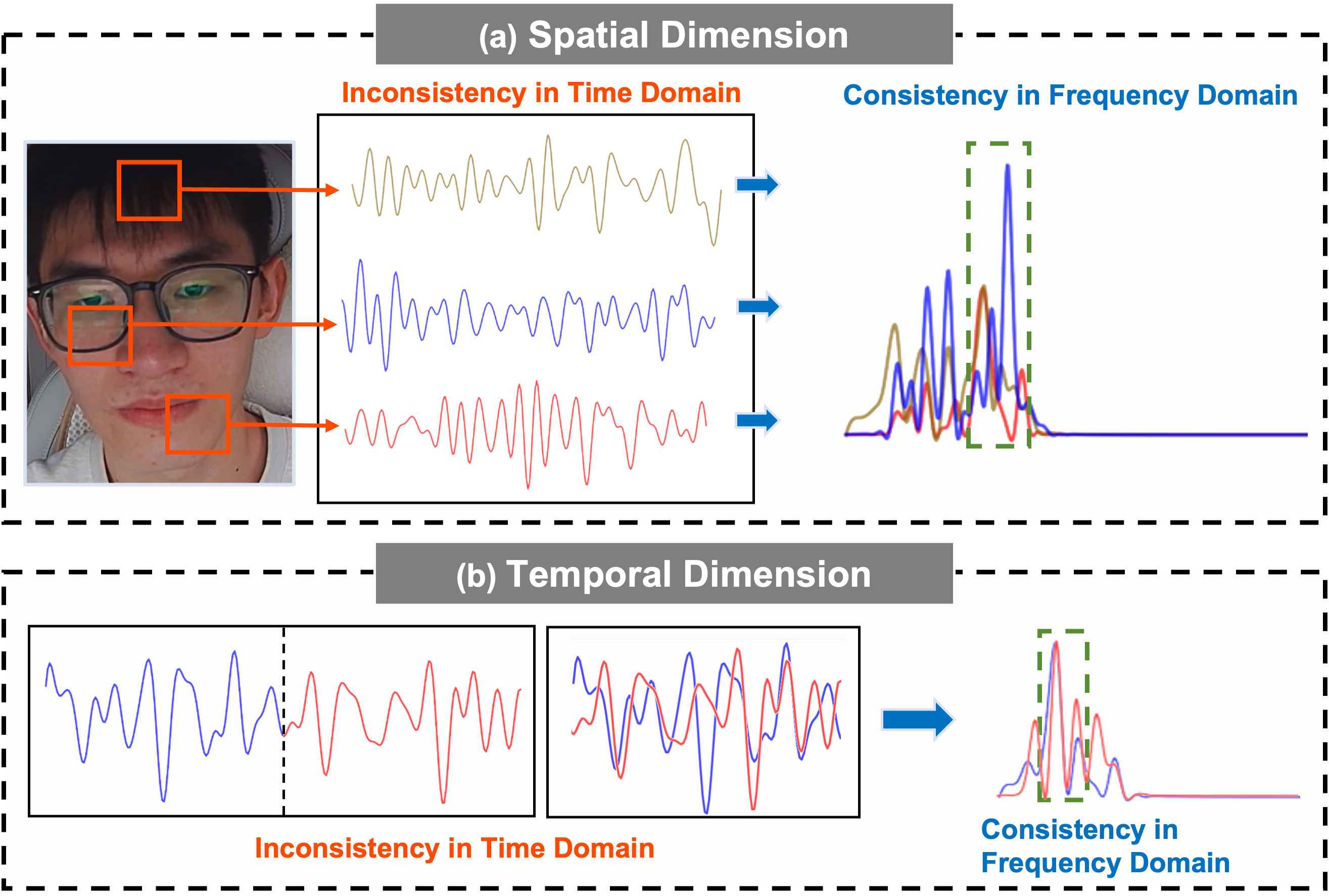}\\
\end{center}
\caption{Illustration of spatio-temporal inconsistency in BVP signals. To better understand the inconsistency, we recover
rPPG signals with the GREEN \citep{verkruysse2008remote} algorithm, which is a reputable classic baseline. (a) indicates that across the same short-time period (8 seconds), various facial regions display comparable PSDs, yet their signal morphology varies. (b) illustrates a continuous BVP signal from the same facial area, where the PSD of two 4-second consecutive segments appears similar, despite differences in signal morphology.}\label{inconsist}
\end{figure}

\subsubsection{Time-domain Inconsistency}
Previous works \citep{wang2023hierarchical,lu2023neuron} in rPPG highlight that enhancing invariant features is key for achieving DG. Notably, semantic spatio-temporal consistency is vital for learning invariant features. As \cite{sun2024} noted, ignoring noise, HR observed in one subject at similar timestamps or facial regions is consistent. Similarly, RR and SpO2 should also adhere to this principle \citep{wang2024physmle}. Numerous studies \citep{li2024bi,xie2024sfda} utilized this concept to strengthen the model's invariant feature learning.  

However, we propose that semantic spatio-temporal consistency should be termed 'frequency-domain spatio-temporal consistency', as time-domain consistency is lacking. For instance, as illustrated in Figure \ref{inconsist}, BVP signals from different facial regions exhibit similar power spectral density (PSD) yet differ in shape and amplitude in the time domain. According to optical and biometrics insights \citep{smith1999pulse,lister2012optical}, variations in facial skin properties and distances from the heart contribute to this. From a temporal viewpoint, factors like respiratory movements \citep{persson1996modulation}, vascular tension in large and microvessels \citep{johnson1986nonthermoregulatory, heistad1973interaction}, and thermoregulation \citep{ivanov2014network} induce short-term oscillations in blood volume. Thus, while consistency acts as a generalizable prior, this inconsistency, largely shaped by individual differences (e.g., skin color, physical condition, respiratory patterns), reflects a variation term of physio-semantic.

These findings have shaped our understanding of Eq. (\ref{eq1}). Firstly, we aimed to reduce the semantically irrelevant noise in both DG and TTA. Furthermore, the semantic part was divided into two: 1. Invariant semantics, referring to optical features linked to facial physiological states (e.g., linear relationships between RoR and SpO2, facial color changes, and BVP); and 2. Individual-specific bias, the impact of personal traits on these relationships. Our study builds on this. Overall, we aim to improve MSSDG by enhancing invariant semantics while reducing instance-specific bias and noises, while promoting instance-specific bias in TTPA for personalization.


\subsection{Prior-based Augmentation}
\label{subsection: Augmentation}
The partial labeling in MSSDG might lead to a shift in the optimization direction of the model towards tasks or domains with more supervised signals \citep{wang2024physmle}. Meanwhile, considering the self-supervised learning manner of TTA, we decided to design a novel data augmentation based on multiple expert priors in rPPG to provide supplementary supervised signals in the form of self-supervision, utilizing the same semantics of the original samples and the augmented data.

\begin{figure}
\begin{center}
\includegraphics[scale=0.1]{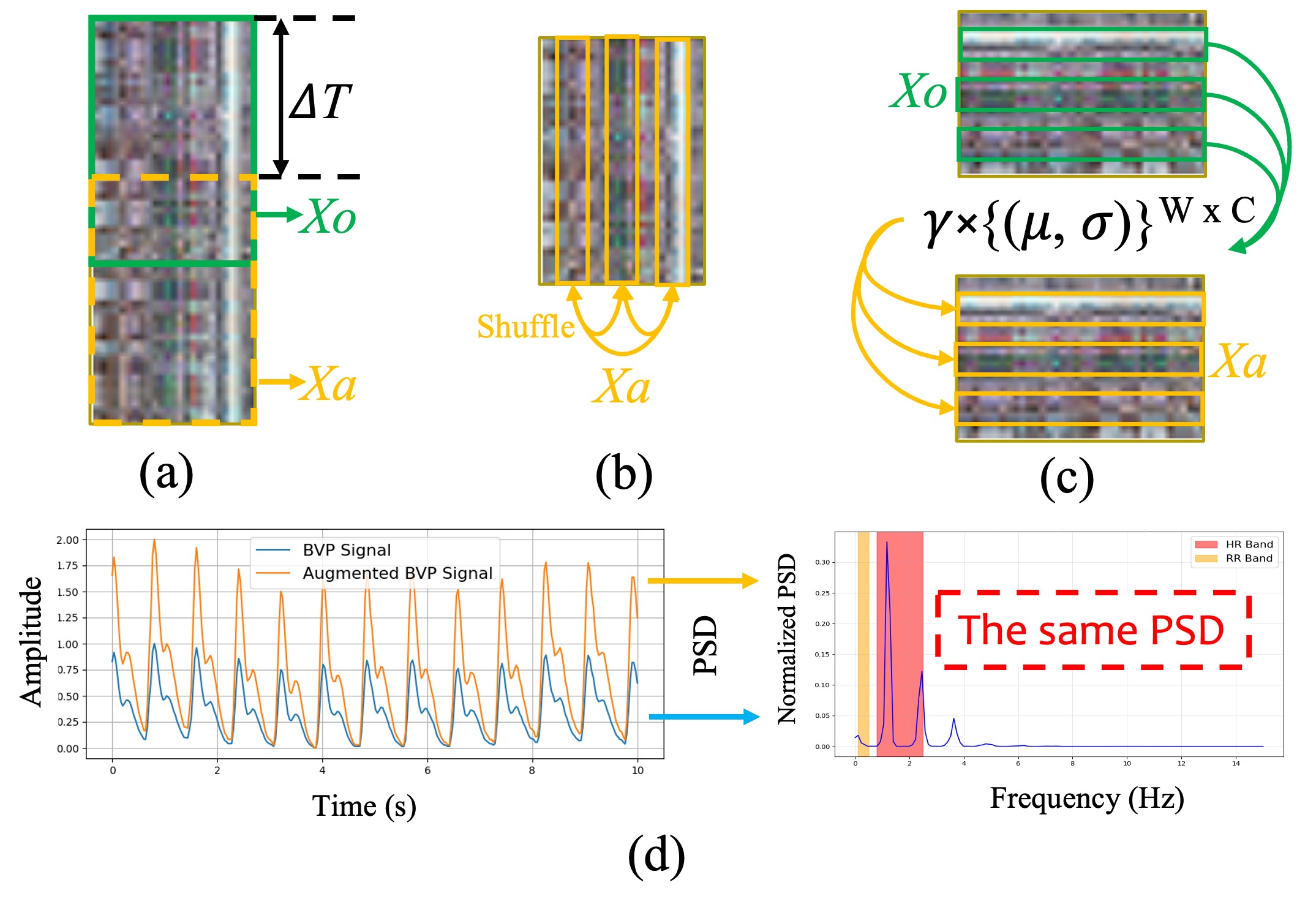}\\
\end{center}
\vspace{-3mm}
\caption{This illustration depicts various data augmentation strategies. Subfigures (a) and (b) represent temporal and spatial augmentations, respectively. In subfigure (a), a sliding window is applied to the original STMap to simulate a time shift, while subfigure (b) illustrates spatial variations by shuffling the rows of the STMap, which correspond to ROIs on the face. Subfigure (c) shows the component scaling augmentation. For this, we compute the mean ($\mu$) and standard deviation ($\sigma$) of the original STMap for each row and channel, then multiply $\mu$ and $\sigma$ by a factor of $\gamma$ to generate the augmented STMap. The concept is further explained in subfigure (d). It is important to note that the PSD of both the original and augmented signals remains unchanged, although the morphology in the time domain is changed.  }\label{aug}
\end{figure}

Firstly, to leverage the physiological signals' spatio-temporal properties, we applied an augmentation strategy over the original STMap $Xo$. In reference to Figure \ref{aug}(a)(b), we began by introducing a random offset variable, denoted as $\Delta T$, which follows a discrete uniform distribution, specifically $\Delta T \sim Uniform(0, 1, 2, \ldots, 30)$. This variable is intended to simulate potential time delays (i.e., 0 to 1 second) and sampling inconsistencies that may occur during the acquisition of physiological signals. The offset $\Delta T$ was then applied to the $T$ dimension of $Xo$. We simulated irregular changes in spatial information by randomly permuting pixels along the $W$ dimension, whose physical meaning was the different face ROIs. This operation involved performing an independent random permutation of the columns at each frame. Specifically, we permute the pixel $Xo^{t, i, j}$ at each time step $t$, where $i$ represents the index along the width dimension and $j$ corresponds to the color channel. The permutation function is denoted as $\Phi(i)$, and the operation can be formally expressed as $Xa^{\tau,\Phi(i),j} \leftarrow Xo^{\tau,i,j}$, where $\Phi$ is a random permutation of the set $\{1, 2, ..., W\}$.

In addition, based on the principle of RoR-based SpO2 estimation, we noticed that, the corresponding SpO2 level should not change if the same scaling was performed over the mean and standard deviation of each color channel of the STMap. To ensure that such scaling will not affect the hidden semantics of other tasks (i.e., HR and RR), we applied the same scaling method to the BVP signal. As shown in Figure \ref{aug}(d), we found that the PSD of the original BVP and the scaled BVP are the same. Therefore, we argued that the component scaling will not change the synsemantic information but provide the model with more variations on the time-domain semantic that may be due to environmental noises (e.g., camera gamma correction, illumination). Specifically, for each row $i$ and channel $j$, we calculated the mean $\mu^{ij}$ and standard deviation $\sigma^{ij}$. Then the scaling value $\gamma \in [0.8, 2.2]$ \citep{chen2023deep} is used to scale the $\mu^i, \sigma^i$ as:

\begin{equation}
 Xa^{[i,:,j]} = \frac{\gamma \times (Xo^{[i,:,j]}-\mu^{ij})}{\sigma^{ij}}+\gamma \times \mu^{ij}.
  \label{eq3}
\end{equation}

\subsection{Robust Shared Representation Learning}
\label{subsection: Shared}

\begin{figure*}
\begin{center}
\includegraphics[scale=0.55]{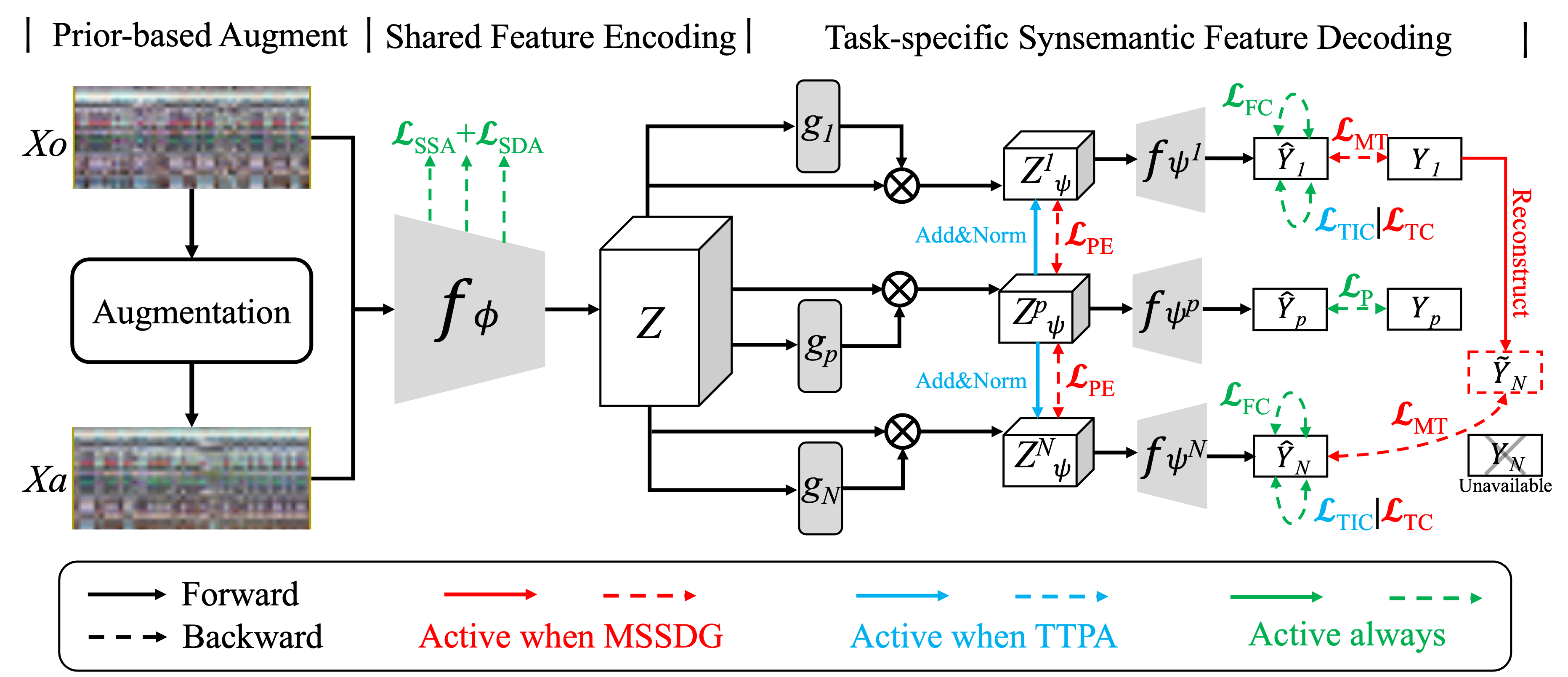}\\
\end{center}
\caption{An overview of the proposed model structure is presented. It consists of three parts: prior-based augmentation, shared feature encoding, and adaptive synsemantic learning. Different operations are activated for various optimization purposes, which are highlighted in different colors. Solid lines represent forward computation, while dashed lines represent reverse optimization. The \textcolor{mssdgred}{RED} font and lines indicate that this operation is only activated during MSSDG, \textcolor{ttpablue}{BLUE} indicates that it is activated during TTPA, and \textcolor{alwaysgreen}{GREEN} indicates that it is always activated. }\label{model}
\end{figure*}

The overall framework of the proposed GAP is shown in Figure \ref{model}. In general, there are two parts after the data augmentation: 1. shared feature encoding; and 2. task-specific synsemantic feature decoding. Firstly, after the data augmentation, we aimed to leverage the shared feature encoder $f_\phi$ to extract the shared large-scale feature for all tasks. However, according to Eq. (\ref{eq1}), we still faced the challenge of semantically irrelevant noise regardless of generalization or personalization, particularly considering that noise (e.g., motion, illumination) in rPPG tasks is usually reflected in low-level large-scale feature structures \citep{wang2023hierarchical}. To help the encoded feature structure be consistently robust to various noises without compromising semantic information, we proposed two forms of regularization.

\subsubsection{Semantic Structure Alignment}
Specifically, given a batch of input pair $(Xo, Xa)$, the encoded representation by each block of the encoder is $\{(Z_o^i,Z_a^i)\}^{N_e}\in\mathbb{R}^{B \times T_i\times W_i\times C_i}$, where $N_e$ is the block number of the encoder. In line with \citep{wang2023hierarchical}, we applied batch normalization and instance normalization to the pairs $(Z_o^i, Z_a^i)$ in order to minimize the intra-channel differences in the representations of all samples. However, we retained the inter-channel differences to adhere to the principles of SpO2 estimation. Next, we transformed $(Z_o^i, Z_a^i)$ into $(\bar{Z}_o^{i}, \bar{Z}_a^{i}) \in \mathbb{R}^{B \times D_i}$ by averaging over the $H_i$ dimension and by flattening the $T_i$ and $C_i$ dimensions to $D_i$. After this transformation, we applied the first regularization, the Semantic Distribution Alignment loss $\mathcal{L}_{SSA}$, which aims to stabilize the representation learning by minimizing the difference of semantic distribution and enhancing the representation structural stability \citep{lu2023neuron} between the original and augmented samples. In detail, for $\bar{Z}_o^{i}$ and $\bar{Z}_a^{i}$, the singular value decomposition (SVD) is performed respectively as follows:  

\begin{align}
    \bar{Z}_o^{i} = U_o^i \cdot \Sigma_o^i \cdot V_o^{i\top}&, \quad \bar{Z}_a^{i} = U_a^i \cdot \Sigma_a^i \cdot V_a^{i\top},  \nonumber\\
      R_{oa}^i = U_o^i \cdot\Sigma_{a}^i\cdot V_o^{i\top}&, \quad R_{ao}^i = U_a^i \cdot\Sigma_{o}^i \cdot V_a^{i\top},  \nonumber\\
    \mathcal{L}_{SSA} &= \sum^{N_e}||R_{oa}^i-R_{ao}^i||_2,
    \label{eq4}
\end{align}

where $U\in\mathbb{R}^{B\times B}$ and $V\in\mathbb{R}^{D_i\times D_i}$ is the left and right singular vector matrix, $\Sigma \in\mathbb{R}^{B\times D_i}$ is the diagonal singular value matrix. By interactively measuring the reconstruction loss after exchanging the singular value matrix of the representations of $Xo$ and $Xa$, $\mathcal{L}_{SSA}$ can quantify whether their distributions and structures are consistent and stable with different noise. 

\subsubsection{Semantic Distribution Alignment}
Although $\mathcal{L}_{SSA}$ can align the local structure in representation space, it does not ensure whether the semantic distribution is also robust to noise. The semantic distribution refers to the direction of the semantic information corresponding to the representations within the batch, whereas semantic structure stabilization does not mean that each of these representations corresponds to a semantic direction. Consequently, inspired by \cite{ahn2024style}, we proposed the Semantic Distribution Alignment loss ($\mathcal{L}_{SDA}$) to ensure that the semantic distribution remains robust under different noise disturbances, through constructing the semantic distribution of samples within a batch and measuring the difference between the original and augmented samples. Specifically, we calculated the covariance matrix $\mathcal{M}^i\in \mathbb{R}^{B\times B}$ for $\bar{Z}_o^{i}$ and $\bar{Z}_a^{i}$. $\mathcal{M}^i$ represents the semantic distribution by assessing correlations among a batch of samples. The whole process was formulated as follows:

\begin{align}
    \mathcal{M}_o^i = \bar{Z}_o^{i} \cdot \bar{Z}_o^{i\top}&, \quad \mathcal{M}_a^i = \bar{Z}_a^{i} \cdot \bar{Z}_a^{i\top}  \nonumber\\
    \mathcal{L}_{SDA} &= \sum^{N_e}||\mathcal{M}_o^i-\mathcal{M}_a^i||_2.
    \label{eq5}
\end{align}

\subsection{Adaptive Synsemantic Learning}
\label{subsection: Synsemantic}
After obtaining the final shared representation $Z$ from $f_\phi$, the GAP was going to perform adaptive computation and optimization depending on the optimization purpose and task. In all, there are $N+1$ estimation tasks, including $N$ physiological estimation tasks and one auxiliary task 
for personal identification. Then, to determine the different semantic information required for each estimation task from the shared synsemantic-mixed feature, we needed to disentangle and reconstruct the feature space before inputting the representation to the task-specific decoders. Following \citep{wang2024physmle}, an element-wise feature gate mechanism $\{g_i\}^{N}_{i=1}$ is applied to adaptively control the importance of low-level shared information. Each $g_i$ shared no parameter and was composed of two neural layers, incorporating a normalization layer and a ReLU activation function. Following traversal through these layers, a Sigmoid function was applied to facilitate element selection via the dot product operation with the original input vector $Z$ to the $\{Z_\psi^i\}^{N+1}_{i=1}$. Next, we will introduce the proposed model in two parts: 1. Synsemantic Modeling and Prior Constraints 2. Individual Difference Disentanglement and Fusion.

\subsubsection{Synsemantic Modeling with Priors}
In this part, we aimed to further extract task-specific semantic information based on the useful features selected for each task. However, under MSSDG, our supervised learning suffers from partial labeling, which in turn leads to the between-task seesaw effect, whereas in the TTPA, we were required to learn without labeling. Therefore, facing the under-labeled scenario in both MSSDG and TTPA, we proposed a prior-driven adaptive self-supervision paradigm. 

In general, we utilized the augmented samples to compare with the original samples to enhance decoders to remain robust to environmental changes while constraining the output with spatio-temporal properties. Specifically, for one estimation task $Y_i$, we decoded $Z^i_\psi$, the corresponding $f^i_\psi$, to $\hat{Y}_i$. Then, the $\hat{Y}_i$ can be further divided to $\hat{Y}a_{i}$ and $\hat{Y}o_{i}$. Given the frequency-domain spatio-temporal consistency among physiological signals, we proposed the $\mathcal{L}_{FC}$ to regularize the distance between $\hat{Y}a_{i}$ and $\hat{Y}o_{i}$. For tasks like HR, RR, and SpO2, since they already directly reflect the frequency-domain values of the corresponding semantic space embedded in the STMap, we directly used the L1 loss; whereas for temporal outputs such as BVP, we needed to first convert the BVP to the frequency domain space before completing the computation of the derivable distance. Given the outputted BVP signals corresponding to $Xo, Xa$, we obtained their PSD and express them as $Q_o, Q_a$. We used the Kullback-Leibler (KL) divergence as the measurement of the frequency distribution distance:

\begin{equation}
\mathcal{L}_{FC}^{BVP} = \frac{1}{N} \sum_{i=1}^{N} Q_{a}^i \cdot \left( \log(Q_{a}^i) - \log(Q_{o}^i) \right).
  \label{eq6}
\end{equation}

Meanwhile, as we mentioned before, time-domain inconsistency also exists in the BVP signal due to individual differences or other domain-related factors (e.g., device, collection way). Previous works usually used the Pearson correlation coefficient as the measurement. However, its point-to-point correspondence calculation makes it insensitive to overall trend shifts that might be caused by time delays. To address it, inspired by \citep{sun2023resolve}, we used the self-similarity matrix to reflect the time-domain characteristics of BVP. Specifically, we used a sliding window of length $s$, with a step size of $1$, to extract the time windows $U = \{u_1,u_2,u_3,...,u_{T-s+1}\}$ from the BVP signals generated by the STMap. By iterating through all elements in $U$ and calculating the cosine similarity between each pair of elements 
$(u_i, u_j)$, we constructed the following self-similarity matrix 
$\mathcal{M} \in \mathbb{R}^{(T-T_i+1)\times(T-T_i+1)}$:

\begin{equation}
\mathcal{M}_{ij} = Sim(u_i,u_j) = \frac{u_i \cdot u_j}{\|u_i\|_2 \times \|u_j\|_2} \label{eq7}
\end{equation}

\begin{figure}
\begin{center}
\includegraphics[scale=0.09]{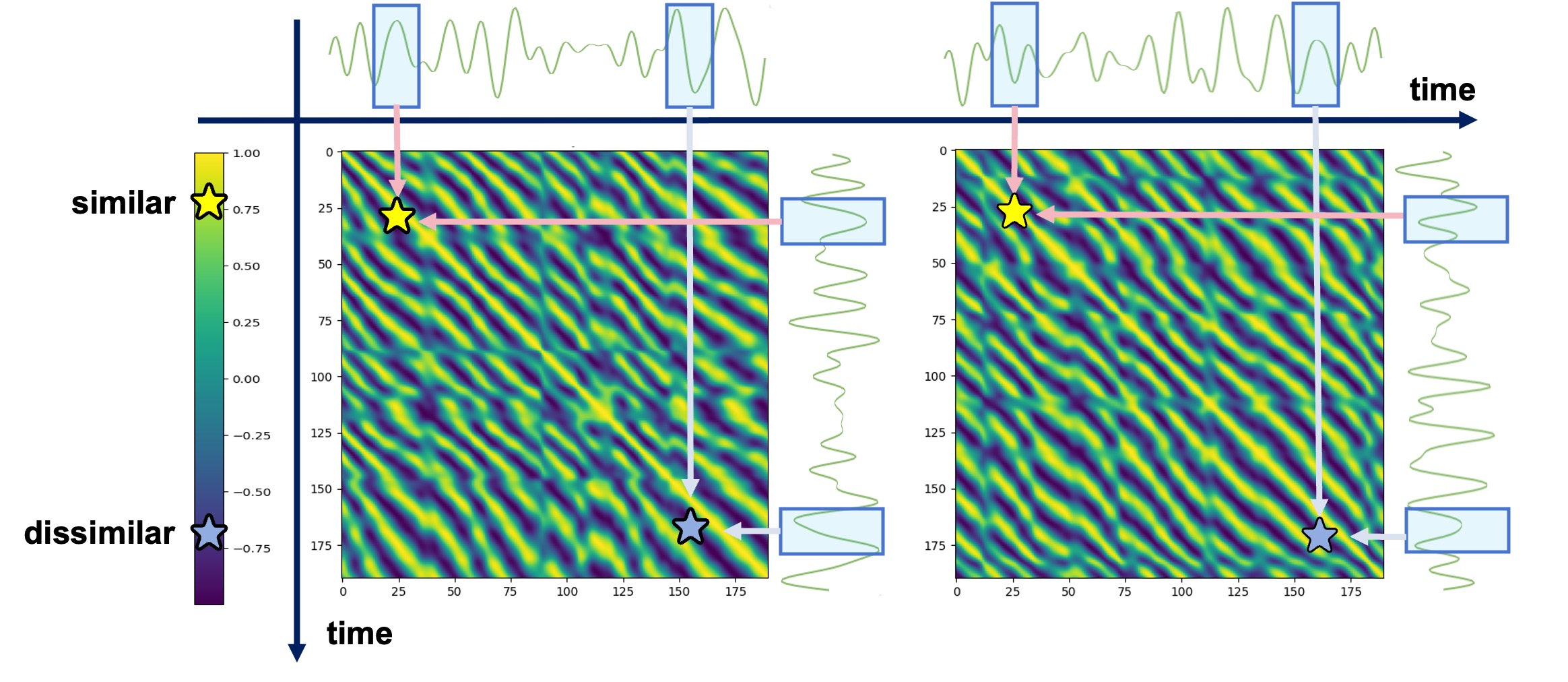}\\
\end{center}
\caption{Visualization of the self-similarity matrix produced from the BVP signals of the STMap prior to and following data augmentation. The intensity of brightness in the image reflects the level of temporal information similarity among the time windows.}\label{ssm}
\end{figure}

Figure \ref{ssm} illustrates that BVP signals can exhibit different morphological traits at identical HR. These variations stem from changes in spatial dimensions, causing notable distinctions in similarity matrices. To effectively capture these local feature variations, we proposed a $\mathcal{L}_{TIC}$ loss that focused on optimizing the dissimilarity between the flattened similarity matrices of the original BVP signals $\mathcal{M}$ and their corresponding augmented counterparts $\mathcal{M}_a$:

\begin{equation}
    \mathcal{L}_{TIC} = 
    \frac{1}{N}\sum^N \frac{\mathcal{M}\cdot \mathcal{M}_a}{\|\mathcal{M}\|_2\times\|\mathcal{M}_a\|_2} \label{eq8}
\end{equation}

Reducing this loss term motivates our model to preserve the unique morphological characteristics of BVP signals rather than averaging vital variations. According to our observations and assertions, this semantic dissimilarity also reflects the specificity of each individual. Thus in the realization of TTPA, this method ensures the model detects subtle differences in signal patterns, leading to more precise and reliable personal physiological measurements. In MSSDG, however, given that we needed to eliminate such instance-level differences, we modified $\mathcal{L}_{TIC}$ to target consistency constraints $\mathcal{L}_{TC}$ on the time domain via $1-\mathcal{L}_{TIC}$.

Last, given the partial labeling in parts of tasks, following our preliminary work \citep{wang2024physmle}, we constructed pseudo-labels based on inter-task correlations from the task with labels to supervise tasks that lack ground truth. For example, if there is a domain that lacks the RR label but has a BVP label, we extracted the RR as a pseudo-label from a specific frequency in its BVP sequence. It is worth noting that since such correlations tend to be unstable, we supervised them with soft constraints, see \citep{wang2024physmle} for details.

\subsubsection{Individual Difference Disentangling and Fusion}
In addition to the main task, in order to more flexibly manipulate individual heterogeneity for different goals, inspired by \citep{han2020leveraging, wang2024rethink}, we explicitly modeled individual attributes in the form of auxiliary tasks. The goal of this auxiliary task was to extract and learn features that can be used to classify each individual based on shared representation $Z$. The extracted individual features are then processed according to the changes in the final application scenario.

Specifically, we first input the shared representation $Z$ to a classification header in the form of a two-layered fully-connected network $f^p_\psi$ to classify the individual $Y_p$ to which the current sample belongs. Note that since it is the MSSDG that usually precedes the TTPA, and the TTPA needs to inherit the model of the MSSDG, we needed to define the output space of the classification header as $N_p+1$ during the MSSDG training, where $N_p$ is the total number of individuals in all source domains. By optimizing a cross-entropy loss $\mathcal{L}_{P}$, we can ensure that $Z^p_\psi$ contains sufficient information about the individual bias of the current sample.

Subsequently, in order to eliminate individual differences in the process of achieving generalizability on the primary tasks, we designed $\mathcal{L}_{PE}$ to ensure orthogonality between the representations $Z^i_{\psi}$ of each primary task and the auxiliary task $Z^p_{\psi}$ that extracted at the middle stage of $f^i_\psi$. In Eq. (\ref{eq9}), since $Z^i_{\psi}$ is not perfectly orthogonal to $Z^p_{\psi}$, we set a constant $\pi$ to constrain $\mathcal{L}_{PE}$ from being too small, and $I$ is the identity matrix. In contrast, for the TTPA in which individual differences need to be reinforced, we performed the dot add and layer normalization operations on each $Z^i_{\psi}$ with $Z^p_{\psi}$ to replace the original $Z^i_{\psi}$. 

\begin{equation}
    \mathcal{L}_{PE} = \frac{1}{N}\sum_{i=1}^N \|Z^p_{\psi} \cdot Z^{i\top}_{\psi}-I\|_2,\mathcal{L}_{PE}>\pi. \label{eq9}
\end{equation}

\subsection{Optimization Goal}
\label{subsection: Optimization}
As shown in Figure \ref{model}, different losses or operations were activated for different purposes. In general, we jointly optimize the network with Eq. (\ref{eq10}), where we incorporated an adaptation factor $p$ \citep{lu2023neuron} to suppress meaningless regularizations during the early iterations. 

\begin{align}
    \mathcal{L}_{U} &= p_1\mathcal{L}_{SSA}+p_2\mathcal{L}_{SDA}+p_3\mathcal{L}_{P}+p_4\mathcal{L}_{FC} \nonumber\\
    \mathcal{L}_{G} &=\sum^N \mathcal{L}_{MT} + p_5\mathcal{L}_{PE}+  p_6\mathcal{L}_{TC}\nonumber\\
    \mathcal{L}_{A} &= p_7\mathcal{L}_{TIC},
    \label{eq10}
\end{align}

where $\mathcal{L}_{MT}$ is the supervised main task losses. We instantiated the L1 loss for HR, RR, and SpO2 estimation, and the negative Pearson’s correlation coefficient loss for BVP estimation. When we aimed to train over the MSSDG protocol, the final loss was the $\mathcal{L}_{U}+\mathcal{L}_{G}$; while the optimization goal turned to $\mathcal{L}_{U}+\mathcal{L}_{A}$ when we needed to achieve personalization.

\section{Experiment}\label{sec3}

\subsection{Dataset}\label{subsec2}

\begin{table*}[h!]
\caption{Summary of the datasets used in this work. \ding{51} means this type of signal is provided in the corresponding dataset, \ding{55} means it is not labeled. Compared to existing public datasets, our proposed HMPC-D provides more comprehensive physiological labels. }\label{t2}
\scriptsize
\setlength{\tabcolsep}{4pt}
\renewcommand{\arraystretch}{1.3}
\centering
\begin{tabular}{l|ccccl}
\toprule
\textbf{Dataset} & \textbf{BVP} & \textbf{HR} & \textbf{SpO2} & \textbf{RR} & \textbf{Challenge} \\ \hline
UBFC-rPPG & \ding{51} & \ding{51} & \ding{55} & \ding{55} & Sunlight and indoor lighting. \\ 
BUAA & \ding{51} & \ding{51} & \ding{55} & \ding{55} & Varying indoor illumination conditions. \\ 
PURE & \ding{51} & \ding{51} & \ding{51} & \ding{55} & Six specific motions. \\ 
VIPL-HR & \ding{55} & \ding{51} & \ding{51} & \ding{55} & Indoor. illumination conditions and motions. \\ 
V4V & \ding{55} & \ding{51} & \ding{55} & \ding{51} & Ten specific tasks. \\ 
HCW & \ding{51} & \ding{51} & \ding{55} & \ding{51} & Varying webcams and cognitive workload. \\ \hline
\textbf{HMPC-D} & \ding{51} & \ding{51} & \ding{51} & \ding{51} & Natural in-vehicle motion and illumination in daytime and night. \\ \bottomrule 
\end{tabular}
\end{table*}

In accordance with \citep{wang2024physmle}, we selected six public datasets to assess the performance of our GAP using both the MSSDG and TTPA protocols. The characteristics of each dataset are summarized in Table \ref{t2}. These datasets encompass a variety of motion, camera, and lighting conditions, providing a holistic evaluation scenario.
Specifically, \textbf{UBFC-rPPG} \citep{bobbia2019unsupervised} includes 42 facial videos shot in both sunlight and indoor lighting. \textbf{BUAA} \citep{xi2020image} focused on evaluating performance under varying illumination conditions. It includes only data with illumination levels of 10 lux or higher. \textbf{PURE} \citep{stricker2014non} contains 60 RGB videos of 10 individuals performing six distinct activities. \textbf{VIPL-HR} \citep{niu2019rhythmnet} presents nine unique scenarios recorded with three RGB cameras, reflecting diverse illumination conditions and movement intensities. \textbf{V4V} \citep{revanur2021first} aims to capture significant fluctuations in physiological indicators, including data from ten specific tasks. \textbf{HCW} \citep{wang2024physmle} collected 48 participants' vital signs under the high cognitive workload. There were various webcams and backgrounds in this dataset.

Notably, the BVP signals and video data within VIPL and V4V are not temporally synchronized, thus, we did not evaluate the BVP and heart rate variability (HRV) estimation performance over it. Additionally, to ensure accurate alignment with the video data, the BVP signals and videos have been resampled to 30 frames per second (fps) using cubic spline interpolation. 

\begin{figure}
\begin{center}
\includegraphics[scale=0.43]{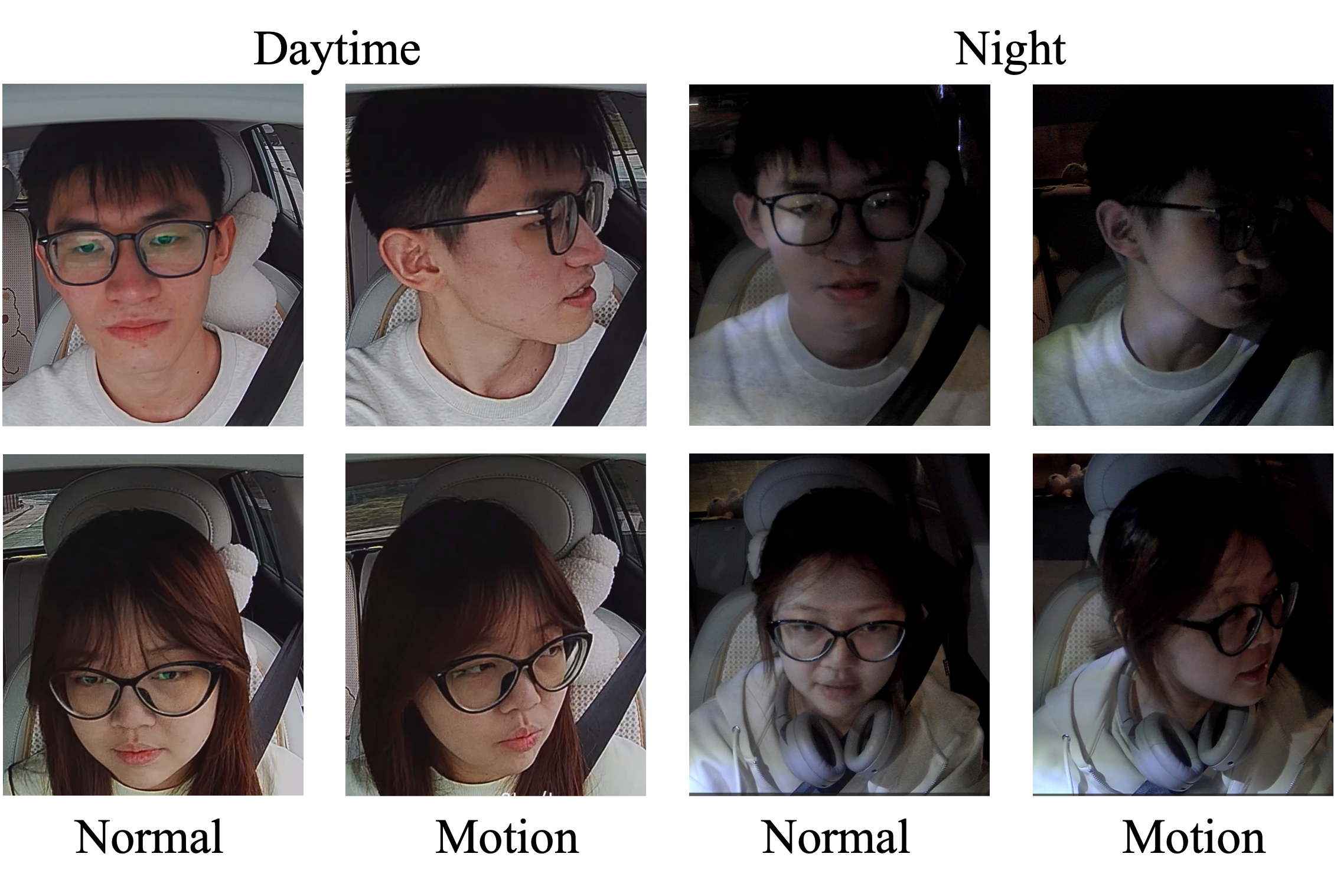}\\
\end{center}
\caption{Example frames in the HMPC-D dataset of the variations in ambient light and head motion during driving.}\label{dataset}
\end{figure}

In addition to the six datasets mentioned earlier, we have also gathered a new dataset, the HIS lab's dataset for Multi-task Physiological measurement with Camera in Driving (HMPC-D), to assess the effectiveness of our dataset in actual commercial deployment settings. While there was previously only one real-world driving dataset (MR-NIRP) \citep{nowara2020near} available in the rPPG domain, it only offers HR labels and lacks RR and SpO2 labels. Furthermore, this dataset involves subjects in the roles of passengers, rather than drivers, which fails to capture the influences of driving maneuvers. 

Consequently, we carried out a driving experiment on an open city road, involving eight participants (four males and four females aged 22 to 28). Each participant was recorded during two driving scenarios: during the day and at night, to examine the effects of varying illumination conditions. Participants were instructed to drive the provided vehicle along a designated route for 15 minutes, and their movements and expressions were not restricted to better reflect the challenges encountered in everyday driving. We captured their frontal images using a 70mai M310 camera, with a resolution of 1920 × 1080 pixels at 30 fps. Example frames are displayed in Figure \ref{dataset}. Ground-truth physiological data for BVP, HR, and SpO2 were obtained with a WS20A finger pulse oximeter, while RR was measured using a respiratory belt transducer at a sampling rate of 100 Hz. The experiment received approval from the Human and Artefacts Research Ethics Committee at the Hong Kong University of Science and Technology (Guangzhou) (protocol number: HKUST(GZ)-HSP-2024-0076). We will make both the raw data (including original video and physiological signals) and the processed data public to the research community.

\subsection{Implementation Details}
\subsubsection{Training Settings}
The entire project was carried out using the PyTorch framework. For data preparation, we adhered to the procedures for STMap generation as detailed in \citep{lu2023neuron}. Each dataset's STMap was sampled every 10 steps within a time window of 256. We instantiated the shared feature encoder with ResNet-18 \citep{he2016deep}, resizing the original STMap from $\mathbb{R}^{25\times 256\times 3}$ to $\mathbb{R}^{64\times 256\times 3}$. For the specific task decoders, we initialized one fully connected layer for HR, SpO2, and RR regression, and utilized four upsample blocks \citep{lu2021dual} for BVP. The experiments were conducted on an Nvidia RTX 3090. 

In the MSSDG, the batch size $B$ was set to 100, with the iteration number $N_{iter}$ being 20000. The hyper-parameter $\pi$ was configured to 0.1. The Adam optimizer was employed for training, with a learning rate of 0.00001. In TTPA, the $B$ was set to 1, and the training will end when one subject's data comes to the end. The optimizer was SGD, with a learning rate of 0.00001. Additionally, we adjusted trade-off parameters $p_4$ through $p_7$ to 0.01, while $p_1$ was set to 0.0001 and $p_2$ to 0.001, aligning with the numeric scale of each loss and their convergence across source domains in MSSDG. 

\begin{table*}[h]
\caption{Comparison Experiment Results of PURE and VIPL-HR on both MSSDG and TTPA Protocol. In this and following tables, we presented the average value of 5-time evaluations. The \textbf{bold} texts show the best result within each column, and the \underline{underlined} texts show the second-best result.}\label{t3}
\scriptsize
\setlength{\tabcolsep}{0.5pt}
\begin{tabular*}{\textwidth}{@{\extracolsep\fill}lccccccccccccc}
\toprule%
& \multicolumn{6}{@{}c@{}}{PURE} & \multicolumn{6}{@{}c@{}}{VIPL-HR} \\\cmidrule{2-7}\cmidrule{8-13}%
& \multicolumn{3}{@{}c@{}}{HR} & \multicolumn{3}{@{}c@{}}{SpO2} & \multicolumn{3}{@{}c@{}}{HR} & \multicolumn{3}{@{}c@{}}{SpO2} \\\cmidrule{2-4}\cmidrule{5-7}\cmidrule{8-10}\cmidrule{11-13}%
Method & MAE↓ & RMSE↓ & p↑ & MAE↓ & RMSE↓ & p↑ & MAE↓ & RMSE↓ & p↑ & MAE↓ & RMSE↓ & p↑ \\
\midrule
\multicolumn{13}{c}{\emph{Baseline methods with MSSDG Protocol}} \\
\midrule
CHROM & 9.79 & 12.76 & 0.37 & — & — & — & 11.44 & 16.97 & 0.28 & — & — & — \\
POS & 9.82 & 13.44 & 0.34 & — & — & — & 14.59 & 21.26 & 0.19 & — & — & — \\
ARM-SpO2 & — & — & — & 3.25 & 3.51 & 0.20 & — & — & — & 18.11 & 20.23 & 0.01 \\
BVPNet & 7.44 & 10.51 & 0.64 & — & — & — & 10.62 & 14.1 & 0.32 & — & — & — \\
PhysFormer++ & 7.16 & 10.14 & 0.65 & — & — & — & 9.91 & 14.05 & 0.37 & — & — & — \\
EfficientPhys & 8.21 & 11.30 & 0.57 & — & — & — & 11.16 & 14.95 & 0.30 & — & — & — \\
rSPO & — & — & — & 3.41 & 3.60 & 0.18 & — & — & — & 14.36 & 15.54 & 0.02 \\
\midrule
\multicolumn{13}{c}{\emph{DG methods with MSSDG Protocol}} \\
\midrule
AD        & 7.52  & 10.79 & 0.6  & 3.15 & 4.26 & 0.19 & 9.81  & 13.99 & 0.37 & 14.78 & 17.98 & 0.04 \\
VREx    & 8.00     & 12.87 & 0.66 & 2.76 & 3.42 & 0.20  & 10.79 & 15.13 & 0.32 & 14.83 & 18.40  & 0.03 \\
GroupDRO  & 8.98  & 10.52 & 0.62 & 3.20  & 4.22 & 0.19 & 15.65 & 17.80  & 0.22 & 16.30  & 21.48 & 0.01 \\
MixStyle & 7.16 & 9.23 & 0.58 & 1.92 & 3.75 & 0.19 & 9.97 & 14.28 & 0.33 & 10.32 & 16.41 & 0.08 \\
NEST    & 6.57  & 10.03 & 0.67 & 3.57 & 4.17 & 0.19 & 10.44 & 14.69 & 0.33 & 15.61 & 18.78 & 0.03 \\
HSRD         & 5.91  & 8.39  &0.71 & 2.80  & 3.98 & 0.20  & 10.04 & 14.33 & 0.36 & 14.83 & 19.41 & 0.06 \\
PhysMLE     & \underline{5.43}  & \underline{7.88} & \textbf{0.77} & \underline{2.19} & \underline{3.23} & \underline{0.22} & \textbf{7.53}  & \textbf{10.29} & \textbf{0.62} & \underline{10.94}  & \underline{14.81} & \underline{0.09}\\
\textbf{GAP-G} & \textbf{5.27} & \textbf{7.57} & \underline{0.74} & \textbf{1.62} & \textbf{2.04} & \textbf{0.31} & \underline{8.16} & \underline{10.67} & \underline{0.60} & \textbf{5.06} & \textbf{8.88} & \textbf{0.13}   \\
\midrule
\multicolumn{13}{c}{\emph{TTA methods with TTPA Protocol}} \\
\midrule
TENT& 6.32 & 9.98 & 0.67 & 2.25 & 2.80 & 0.25 & 9.35 & 12.02 & 0.54 & 6.43 & 10.21 & 0.09 \\
SAR& 6.01 & 9.15 & 0.68 & 2.10 & 2.63 & 0.26 & 9.41 & 12.18 & 0.52 & 7.07 & 10.86 & 0.07 \\
TTT++& 5.32 & 7.71 & 0.71 & 1.90 & 2.38 & 0.28 & 8.05 & 10.12 & 0.61 & 4.93 & 8.42 & 0.14 \\
EATA& 5.39 & 7.87 & 0.70 & 1.88 & 2.27 & 0.29 & 7.92 & 10.04 & 0.62 & 5.03 & 8.51 & 0.13 \\
AdaODM& 5.83 & 8.02 & 0.69 &  1.90 & 2.11 & 0.29 & \underline{7.77} & \underline{9.91} & \underline{0.63} & 4.91 & 8.41 & 0.14 \\
Bi-TTA& \underline{4.89} & \underline{6.31} & \underline{0.76} & \underline{1.83} & \underline{2.03} & \underline{0.31} & 8.33 & 10.87 & 0.61 &  \underline{4.80} & \underline{8.30} & \underline{0.15} \\
\textbf{GAP-P} & \textbf{4.51} & \textbf{5.82} & \textbf{0.80} & \textbf{1.23} & \textbf{1.68} & \textbf{0.35} & \textbf{7.02} & \textbf{9.10} & \textbf{0.71} & \textbf{4.52} & \textbf{6.38} & \textbf{0.17} \\

\botrule
\end{tabular*}
\end{table*}

\subsubsection{Experimental Settings}

We set two protocols: (1) MSSDG protocol; this protocol was implemented in a cross-domain evaluation scenario by dividing all datasets into two groups: one dataset (target domain) was invisible in training and was used for testing the model that was trained on the other datasets (source domain). Specifically, in our training on source domains, we follow the method outlined in \citep{liang2020we}, randomly splitting the source data into a 9:1 ratio. 90\% data is used to train the source model, while the remaining 10\% is utilized for validation and hyper-parameter selection. Then, the trained model will be evaluated on the target domain. (2) TTPA protocol, which begins after the training of MSSDG. After training the model from source domains, its training and evaluation on the target domain were individual-specific. Specifically, we needed to perform an adaptation for each individual in the target domain in chronological order. After each completed adaptation, the model needed to perform forward inference on the current sample. The repeated adaptation on the same sample was not allowed. It is worth noting that since TTPA focused more on personalization, all users can only use the same post-DG model parameters for adaptation and cannot learn successively across users. Subsequently, referring to our method design, the proposal GAP could be classified into two variants: GAP-G for MSSDG and GAP-P for TTPA.

We adopted evaluation metrics from prior studies \citep{lu2023neuron, wang2023hierarchical}, including mean absolute error (MAE), root mean square error (RMSE), and Pearson’s correlation coefficient (p) to evaluate estimated physiological indicators. Additionally, for measuring HRV, we employed normalized low frequency (LFnu), normalized high frequency (HFnu), and the LF/HF ratio (the ratio between low-frequency and high-frequency power). We performed five evaluations using five different random seeds, and the average performance was reported in the subsequent sections.

\begin{table*}[t]
\caption{Comparison Experiment Results of V4V and HCW on both MSSDG and TTPA Protocol.}\label{t4}
\setlength{\tabcolsep}{1mm}
\centering
\scriptsize
\setlength{\tabcolsep}{0.5pt}
\begin{tabular*}{\textwidth}{@{\extracolsep\fill}lccccccccccccc}
\toprule%
& \multicolumn{6}{@{}c@{}}{V4V} & \multicolumn{6}{@{}c@{}}{HCW} \\
\cmidrule{2-7}\cmidrule{8-13}%
& \multicolumn{3}{@{}c@{}}{HR} & \multicolumn{3}{@{}c@{}}{RR} & \multicolumn{3}{@{}c@{}}{HR} & \multicolumn{3}{@{}c@{}}{RR} \\\cmidrule{2-4}\cmidrule{5-7}\cmidrule{8-10}\cmidrule{11-13}%
Method & MAE↓ & RMSE↓ & p↑ & MAE↓ & RMSE↓ & p↑ & MAE↓ & RMSE↓ & p↑ & MAE↓ & RMSE↓ & p↑ \\
\midrule
\multicolumn{13}{c}{\emph{Baseline methods with MSSDG Protocol}} \\
\midrule
CHROM       & 14.92         & 19.22          & 0.08 & —   & —   & —   & 14.66 & 19.81 & 0.05 & —    & —    & —   \\
 POS         & 12.67         & 18.11          & 0.06  & —   & —   & —   & 12.37 & 17.61 & 0.11 & —    & —    & —   \\
 ARM-RR            & —    & —    & —   & 6.33 & 6.90 & 0.01  & —    & —    & —   & 8.02  & 8.66 & 0.01 \\
BVPNet     & 11.16 & 15.10 & 0.31   & —   & —   & —       & 9.52  & 12.01 & 0.57   & —    & —    & —   \\
PhysFormer++  & 9.53  & 12.96 & 0.37   & —   & —   & —       & 8.12  & 10.05 & 0.51   & —    & —    & —   \\
EfficientPhys  & 11.87 & 16.02 & 0.27   & —   & —   & —       & 10.02 & 11.79 & 0.34   & —    & —    & —   \\
MTTS-CAN      & 12.06 & 15.01 & 0.11   & 6.44 & 7.80 & 0.01     & 13.38 & 15.21 & 0.21   & 9.33  & 15.86 & 0.01   \\
BigSmall     & 11.88 & 14.60 & 0.29   & 6.36 & 7.68 & 0.02     & 12.05 & 14.11 & 0.25   & 8.42  & 14.30 & 0.01   \\
\midrule
\multicolumn{13}{c}{\emph{DG methods with MSSDG Protocol}} \\
\midrule
AD      & 10.91 & 13.72 & 0.31   & 3.06 & 3.83 & 0.14     & 9.86 & 11.14 & 0.41   & 7.54  & 13.84 & 0.04 \\
VREx      & 10.96 & 13.78 & 0.31   & 3.23 & 4.23 & 0.13     & 8.23 & 10.93 & 0.49   & 7.44  & 12.56 & 0.06 \\
GroupDRO  & 10.58 & 12.92 & 0.32   & 3.98 & 4.47 & 0.12     & 9.11  & 11.39 & 0.44   & 8.28  & 13.17 & 0.03 \\
Mixstyle     & 11.76 & 14.47 & 0.30   & 3.48 & 4.11 & 0.13     & 9.06  & 10.80 & 0.44   & 7.83  & 3.28 & 0.07 \\
NEST   & 9.62  & 11.05 & 0.38   & 3.46 & 4.50 & 0.12     & 9.76 & 11.35 & 0.41   & 7.31  & 12.59 & 0.04 \\
HSRD & 9.18  & 10.83 & 0.42   & 3.11 & 4.01 & 0.13     & 9.35  & \textbf{10.02} & 0.47   & 7.28  & 12.92 & 0.04 \\
PhysMLE    & \underline{8.12}  & \underline{10.56} & \underline{0.51}   & \underline{3.02} & \underline{3.66} & \underline{0.14}     & \textbf{7.13}  & \underline{10.20}  & \textbf{0.61}   & \underline{5.90}  & \underline{8.84}  & \underline{0.10} \\
\textbf{GAP-G}& \textbf{7.60} & \textbf{9.82} & \textbf{0.58} &\textbf{2.84} & \textbf{3.13}& \textbf{0.19} & \underline{8.07} & 10.51 & \underline{0.55} & \textbf{2.08} & \textbf{2.58} & \textbf{0.31} \\
\midrule
\multicolumn{13}{c}{\emph{TTA methods with TTPA Protocol}} \\
\midrule
TENT& 8.02 & 10.33 & 0.55 & 3.09 & 3.59 & 0.15 & 8.20 & 10.86 & 0.52 & 3.11 & 3.77 & 0.21 \\
SAR& 7.92 & 10.14 & 0.56 & 3.01 & 3.48 & 0.15 & 8.03 & 10.66 & 0.54 & 3.02 & 3.42 & 0.23 \\
TTT++& 7.97 & 10.35 & 0.55 & 2.96 & 3.26 & 0.16 & 7.92& 10.08 & 0.56 & 2.87 & 3.03 & 0.25 \\
EATA& 7.56 & 9.83 & 0.58 & 2.79 & 3.20 &0.18 & 8.11 & 10.92 & 0.53 & 2.36 & 2.95 & 0.28 \\
AdaODM& 7.44 & 9.62 & \underline{0.59} & 2.68 & 3.16 & 0.18 & 7.99 & 10.14 & 0.56  & \underline{2.00} & \underline{2.36} & \underline{0.31} \\
Bi-TTA& \underline{7.38} & \underline{9.57} & \underline{0.59} & \underline{2.61} & \underline{3.09} & \underline{0.19} & \underline{7.88} & \underline{10.06} & \underline{0.57}& 2.11 & 2.64 & 0.30 \\
\textbf{GAP-P}    & \textbf{7.23}  & \textbf{9.21} & \textbf{0.61}   & \textbf{2.05} & \textbf{2.73} & \textbf{0.21}     & \textbf{7.30}& \textbf{9.29} & \textbf{0.59}   & \textbf{1.90}  & \textbf{2.11}  & \textbf{0.33} \\
\bottomrule 
\end{tabular*}
\end{table*}

\subsubsection{Baselines}

We evaluated the performance of GAP across different datasets under MSSDG and TTPA protocols. Based on the MSSDG benchmark we built \citep{wang2024physmle}, we selected some representative traditional single-task methods (i.e., CHROM \citep{de2013robust}, POS \citep{wang2016algorithmic}, and ARM \citep{tarassenko2014non}), deep single-task learning-based methods (i.e., BVPNet \citep{das2021bvpnet}, PhysFormer++ \citep{yu2023physformer++}, EfficientPhys \citep{liu2023efficientphys}, rSPO \citep{akamatsu2023blood} ), deep multi-task learning-based methods (i.e., MTTS-CAN \citep{liu2020multi}, and BigSmall \citep{narayanswamy2024bigsmall}), and DG-based methods (i.e., AD \citep{ganin2015unsupervised}, GroupDRO \citep{parascandolo2020learning}, VREx \citep{krueger2021out}, Mixstyle \citep{zhou2024mixstyle}, NEST \citep{lu2023neuron}, HSRD \citep{wang2023hierarchical}) for MSSDG training and evaluation. Meanwhile, several TTA-based methods (i.e., TENT \citep{wang2020tent}, SAR \citep{niu2023towards}, TTT++ \citep{liu2021ttt++}, EATA \citep{niu2022efficient}, AdaODM \citep{zhang2023adaptive}, Bi-TTA \citep{li2024bi} ) were used to assess the performance on the TTPA protocol.

Notably, all DG methods, apart from PhysMLE, were developed using a multi-task variant featuring a single ResNet-18 backbone and four task estimation heads similar to GAP. For a fair comparison, all TTA baselines were built based on GAP-G. Additionally, we removed the action units prediction head of BigSmall \citep{narayanswamy2024bigsmall} and the GRU mechanism from Rhythmnet \citep{niu2019rhythmnet} to avoid the negative impact of the extra heads. For the previous TTA methods designed for classification (such as TENT, SAR, TTT++, EATA, and AdaODM), we divided the ideal HR value range into 24 equal intervals, while we separated SpO2 and RR into 10 intervals each. We then added three classification heads to GAP, each consisting of a single-layer neural network, to determine which interval the current sample falls into for these three tasks. This approach adopts these baselines to rPPG tasks.

\subsection{Performance on Public Datasets}
\subsubsection{MSSDG protocol}
The results comparing our proposal across six public datasets are presented in Tables \ref{t3}, \ref{t4}. Notably, under the MSSDG protocol, the GAP-G model excels in the HR task on the PURE dataset, achieving a leading MAE of 5.27 and RMSE of 7.57. This represents an improvement of approximately 10\% in MAE and 12\% in RMSE when compared to the second-best model, PhysMLE. For the VIPL-HR dataset, even though PhysMLE surpasses our proposal in HR estimation, GAP-G still performs admirably with an MAE of 8.16 and RMSE of 10.67, marking around a 9\% and 11\% enhancement over HSRD. Specifically, in the SpO2 task, GAP-G significantly outperforms PhysMLE across all metrics, showing improvements of about 9\% in MAE and 42\% in p. In contrast to PhysMLE, which has a larger number of parameters and requires extensive hyper-parameter tuning (the total number of parameters of GAP is slightly higher than ResNet18, but only two-thirds of PhysMLE), our model provides balanced performance with a simpler architecture and training process. The superiority of 'weak' tasks, such as SpO2, further emphasizes that our prior-based augmentation strategy can effectively mitigate the partial labeling in these tasks.

In the HR task for the V4V dataset, the GAP-G model also stands out, surpassing other DG methods, with an MAE of 7.60 and an RMSE of 9.82, especially in comparison to PhysMLE, which shows improvements of approximately 6\% in MAE and 7\% in RMSE. However, in the HR task for the HCW dataset, PhysMLE outperforms GAP-G with an MAE of 7.13 and RMSE of 10.20, indicating its superiority for this particular task. Nonetheless, GAP-G still demonstrates commendable performance in the RR task, achieving an average improvement of around 9\% over other methods. Considering GAP-G's performance on VIPL-HR, we suggest that higher model complexity offers a significant advantage in HR estimation amid unstable motion or recording conditions.
\begin{table*}[]
\setlength{\tabcolsep}{0.6mm}
\caption{HRV and HR estimation results on MSSDG protocol.}
\label{t5}
\centering
\scriptsize
\begin{tabular}{clcccccccccccc}
\toprule
 &   & \multicolumn{3}{c}{LFnu}   & \multicolumn{3}{c}{HFnu}   & \multicolumn{3}{c}{LF/HF} & \multicolumn{3}{c}{HR-(bpm)}     \\
\cmidrule(lr){3-5} \cmidrule(lr){6-8} 
\cmidrule(lr){9-11} \cmidrule(lr){12-14}        
{Target}&{Method}  & MAE↓    & RMSE↓   & \quad p↑ \quad      & MAE↓    & RMSE↓   & \quad p↑ \quad     & MAE↓    & RMSE↓   & \quad p↑ \quad     & MAE↓    & RMSE↓   & \quad p↑ \quad     \\
\midrule 
UBFC & CHROM    & 0.222 & 0.281 & 0.070 & 0.222 & 0.281 & 0.070 & 0.671 & 1.054 & 0.105 & 7.229    & 8.922 & 0.512 \\
& POS      & 0.236 & 0.286 & 0.136 & 0.236 & 0.286 & 0.136 & 0.652 & 0.954 & 0.135 & 7.354    & 8.040  & 0.492 \\
& PhysFormer++ & 0.084 & 0.113 & 0.182 & 0.084 & 0.113 & 0.182 & 0.323 & 0.365 & 0.306 & 5.635    & 7.258  & 0.769 \\
& HSRD       & 0.173 & 0.267 & 0.166 & 0.173 & 0.267 & 0.166 & 0.321 & 0.558 & 0.264 & 6.176    & 7.344  & 0.672 \\
& PhysMLE  & 0.063 & 0.082 & 0.201 & 0.063 & 0.082 & 0.201 & 0.233 & 0.301 & 0.315 & 4.927    & 5.746  & 0.831  \\
& Bi-TTA  & 0.058 & 0.072 & 0.241 & 0.058 & 0.072 & 0.241 & 0.193 & 0.275 & 0.323 & 4.416 & 6.355 & 0.873  \\
& \textbf{GAP-G}  & 0.059 & 0.075 & 0.227 & 0.059 & 0.075 & 0.227 & 0.210  & 0.272 & 0.321& 4.537  & 6.561 & 0.871 \\
& \textbf{GAP-P}  & \textbf{0.051} & \textbf{0.073} & \textbf{0.259} & \textbf{0.051} & \textbf{0.073} & \textbf{0.259} & \textbf{0.191} & \textbf{0.272} & \textbf{0.327} & \textbf{4.399} & \textbf{6.139} & \textbf{0.885}  \\
  \midrule                     
BUAA & CHROM    & 0.379 & 0.324 & 0.068 & 0.379 & 0.324 & 0.068 & 0.681 & 0.884 & 0.072 & 6.093    & 8.294 & 0.517 \\
& POS      & 0.320 & 0.376 & 0.096 & 0.320 & 0.376 & 0.096 & 0.628 & 0.842 & 0.113 & 5.041    & 7.120  & 0.637 \\
& PhysFormer++ & 0.201 & 0.394 & 0.289 & 0.201 & 0.394 & 0.289 & 0.599 & 0.926 & 0.282 & 4.021    & 8.711  & 0.790 \\
& HSRD       & 0.189 & 0.172 & 0.313 & 0.189 & 0.172 & 0.313 & 0.583 & 0.697 & 0.320 & 3.073    & 6.912  & 0.866 \\
& PhysMLE  & 0.148 & 0.169 & 0.291 & 0.148 & 0.169 & 0.291 & 0.562 & 0.698 & 0.299 & 3.456    & 7.038  & 0.834 \\
& Bi-TTA          & 0.136 & 0.142 & 0.321 & 0.136 & 0.142 & 0.321 & 0.514 & 0.617 & 0.322 & 2.736 & 4.046 & 0.789 \\
& \textbf{GAP-G}          & 0.135 & 0.158 & 0.304 & 0.135 & 0.158 & 0.304 & 0.520  & 0.621 & 0.310 & 2.891  & 4.073  & 0.778 \\
& \textbf{GAP-P}          & \textbf{0.124} & \textbf{0.136} & \textbf{0.344} & \textbf{0.124} & \textbf{0.136} & \textbf{0.344} & \textbf{0.503} & \textbf{0.601} & \textbf{0.339} & \textbf{2.682} & \textbf{3.724} & \textbf{0.792} \\
\bottomrule 
\end{tabular}
\vspace{-2mm}
\end{table*}

\subsubsection{TTPA protocol}
In this initial TTPA work on the MTL rPPG task, we showcase the superiority of our proposal by contrasting it with various classical and latest TTA methods in this field. Notably, GAP-P consistently outperforms all other methods across target domains. For instance, when evaluated on the VIPL-HR dataset, GAP-P achieves an MAE of 7.02 and an RMSE of 9.10 in the HR task, reflecting improvements of approximately 13\% and 12\%, respectively, outperforming the second-best model. In the RR task, GAP-P leads with an MAE of 2.73 and RMSE of 3.13, surpassing the less effective TTA model by about 10\% and 9\%, showcasing its adaptability across diverse datasets and tasks. Unlike Bi-TTA, which relies solely on consistency for self-supervised learning in the target domain, our approach emphasizes user-level personalized adaptation while integrating both consistency and inconsistency. This reinforces the validity of our prior work in further disentangling semantics into invariant semantics and individual-specific bias. Moreover, both Bi-TTA and GAP-P, as methods tailored for rPPG, demonstrate superior performance compared to other generalized TTA approaches that are initially designed for classification, highlighting the necessity of specialized methods for specific tasks.

\subsubsection{HRV evaluation}
Lastly, as shown in Table \ref{t5}, we present the HRV estimation results from the UBFC and BUAA datasets, which exclusively provide BVP signal labels. We evaluated the HRV index (LFnu, HFnu, and LF/HF) as well as HR (bpm) to assess the quality of the predicted BVP signal. Generally, the observed performance variations align with the conclusions drawn in Tables \ref{t3} and \ref{t4}. Performance improves progressively, starting from traditional methods (like CHROM and POS) to deep learning models. Our proposals consistently outperformed in all indicators, with both MAE and RMSE at lower levels. This indicates that the model produces a smoother BVP signal with reduced noise interference. Furthermore, the accuracy of LFnu and HFnu suggests that the model is more effective at capturing the low and high-frequency components of the BVP signal, benefiting applications in emotion recognition and healthcare.

\begin{table*}[h]
\centering
\caption{Performance comparison over the HMPC-D based on both MSSDG and TTPA protocol. Evaluations and adaptations on day, night, and full datasets are shown separately. }\label{t6}
\scriptsize
\setlength{\tabcolsep}{0.5pt}
\begin{tabular*}{\textwidth}{@{\extracolsep{\fill}}lccccccccccc}
\toprule
& & & CHROM & ARM-RR & ARM-SpO2 & MTTS-CAN & BigSmall & HSRD & PhysMLE & GAP-G & GAP-P \\
\midrule
\multirow{2}{*}{Day} & HR & RMSE & 33.34 & — & — & 19.26 & 18.29 & 15.33 & 11.91 & 12.09 & 11.34 \\
& & p & 0.01 & — & — & 0.05 & 0.06 & 0.11 & 0.21 & 0.20 & 0.22 \\
\cmidrule(l){2-12}
\multirow{2}{*}{} & RR & RMSE & — & 18.24 & — & 8.92 & 9.03 & 7.33 & 7.60 & 6.46 & 5.16 \\
& & p & — & 0.01 & — & 0.03 & 0.03 & 0.04 & 0.04 & 0.07 & 0.09 \\
\cmidrule(l){2-12}
\multirow{2}{*}{} & SpO2 & RMSE & — & — & 14.38 & — & — & 10.96 & 10.17 & 9.01 & 8.26 \\
& & p & — & — & 0.03 & — & — & 0.08 & 0.09 & 0.12 & 0.14 \\
\midrule
\multirow{2}{*}{Night} & HR & RMSE & 30.29 & — & — & 15.90 & 15.32 & 12.28 & 12.46 & 9.90 & 7.27 \\
& & p & 0.02 & — & — & 0.12 & 0.12 & 0.17 & 0.16 & 0.24 & 0.38 \\
\cmidrule(l){2-12}
\multirow{2}{*}{} & RR & RMSE & — & 15.08 & — & 7.23 & 7.52 & 6.31 & 6.23 & 5.95 & 4.71 \\
& & p & — & 0.01 & — & 0.05 & 0.05 & 0.09 & 0.09 & 0.10 & 0.13 \\
\cmidrule(l){2-12}
\multirow{2}{*}{} & SpO2 & RMSE & — & — & 13.04 & — & — & 10.35 & 9.68 & 9.07 & 7.66 \\
& & p & — & — & 0.05 & — & — & 0.10 & 0.11 & 0.11 & 0.14 \\
\midrule
\multirow{2}{*}{All} & HR & RMSE & 31.95 & — & — & 17.85 & 17.24 & 13.29 & 11.16 & 10.87 & 8.79 \\
& & p & 0.01 & — & — & 0.09 & 0.10 & 0.15 & 0.20 & 0.22 & 0.28 \\
\cmidrule(l){2-12}
\multirow{2}{*}{} & RR & RMSE & — & 16.75 & — & 8.07 & 8.26 & 6.88 & 6.02 & 5.35 & 5.04 \\
& & p & — & 0.01 & — & 0.05 & 0.04 & 0.09 & 0.10 & 0.11 & 0.12 \\
\cmidrule(l){2-12}
\multirow{2}{*}{} & SpO2 & RMSE & — & — & 13.52 & — & — & 10.49 & 9.88 & 9.02 & 7.97 \\
& & p & — & — & 0.04 & — & — & 0.09 & 0.10 & 0.11 & 0.13 \\
\bottomrule
\end{tabular*}
\end{table*}

\subsection{Computational Cost}

\begin{table}[h]
\centering
\caption{Model Complexity and Inference Runtime. Param(M) is the number of model parameters in millions, FLOPs(G) refers to the number of floating point operations in gigaflops, and Runtime(ms) is the average per-sample inference time in milliseconds.}\label{cost}
\scriptsize
\setlength{\tabcolsep}{4pt}  
\begin{tabular}{lccc}
\toprule
Method       & Param(M) & FLOPs(G) & Runtime(ms) \\
\midrule
PhysMLE      & 24.82     & 34.57     & 1780         \\
GAP-G        & 29.53     &  3.37     &   56         \\
\midrule
TENT         & 29.54     &  3.37     &  804         \\
Bi-TTA       & 29.54     &  3.37     & 1586         \\
GAP-P        & 29.53     &  3.37     & 1240         \\
\bottomrule
\end{tabular}
\end{table}

To verify the performance of our proposal in terms of computational cost and real-time performance, we compare the model computational complexity and inference time when inputting a sample with the latest multi-task MSSDG method (PhysMLE) and multiple TTA methods (TENT, Bi-TTA) on our own server. We evaluated the number of model parameters (Param(M)), the computational complexity of the model (FLOPs(G)), and the reasoning time (Runtime(ms)). The results are displayed in Table \ref{cost}. It is worth noting that under the TTPA protocol, the runtime we measured includes one backward gradient update and two forward processes.

First of all, the GAP framework has significantly improved in terms of computational complexity and inference time compared with PhysMLE. In particular, the significant reduction in inference time ensures that the GAP framework has excellent deployment and real-time computing potential without considering TTPA. Subsequently, we also evaluated three TTPA methods based on GAP-G. As expected, the runtime has increased significantly compared with that under the MSSDG protocol. Although TENT achieves the minimum runtime, as shown in Table \ref{t3} and Table \ref{t4}, TENT does not improve the performance on the basis of MSSDG; instead, it leads to performance degradation. Compared with Bi-TTA, which is more competitive in estimation performance, our GAP-P has a notable advantage in runtime. Nevertheless, it should be noted that there is still room for improvement in the runtime. Factors that may be further optimized include the computing platform and code construction, etc.

\begin{figure*}
\begin{center}
\includegraphics[scale=0.11]{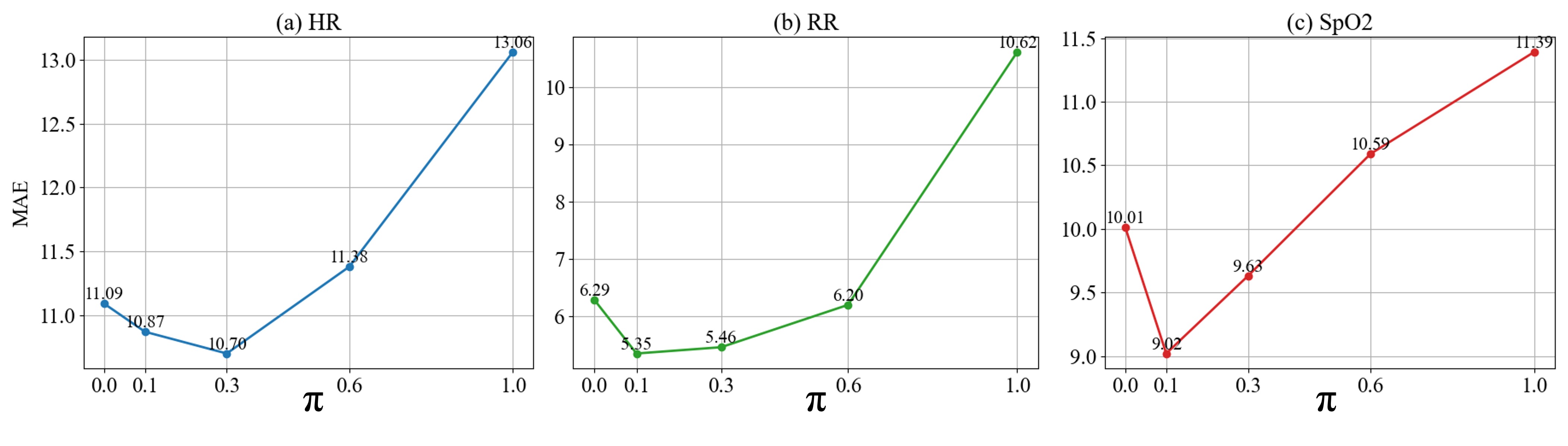}\\
\end{center}
\vspace{-3mm}
\caption{MAE of adopting different hyper-parameter $\pi$ over GAP-G in the HMPC-D.}\label{parameter}
\vspace{-2mm}
\end{figure*}
\subsection{Performance on Driving Scenario}

\subsubsection{Evaluation of Generalization}
Table \ref{t6} presents the performance comparison of various models on the HMPC-D dataset across different physiological indicators (HR, RR, SpO2) during day, night, and overall scenarios. Firstly, during the day, DL models such as MTTS-CAN and BigSmall achieved RMSE values of 19.26 and 18.29, respectively, indicating moderately better performance compared to classic models (CHROM, ARM). Further, DG models (HSRD, PhysMLE, GAP-G) achieved improvements over the DL-based models, where GAP-G consistently outperformed previous DG methods. It aligned with our findings from the six public datasets. 

However, at night, the errors of all methods were slightly lower, which was different from common sense. We argue there might be two reasons: (1) Although the light at night is fainter compared to daytime, considering the constant light generated by the interfaces in the car and the stability of the city street light gaps, the light conditions in the cabin at night are actually relatively faint but more stable compared to daytime; (2) During the daytime in addition to the natural light conditions affected by weather, factors such as shade, relative position of the sun can pose challenges in achieving generalizability. Overall, the error of all models on HMPC-D was greater compared to the previous six public indoor datasets, suggesting that in-vehicle physiological estimation still requires further investigation.

\subsubsection{Parameter Sensitivity Test}

To investigate the impact of the hyperparameter $\pi$ in Eq.(\ref{eq9}) on model performance, we conducted a sensitivity analysis across three physiological tasks. As $\pi$ regulates the minimum strength of the orthogonality constraint between the individual-specific representation $Z^p\_{\psi}$ and each task-specific representation $Z^i\_{\psi}$, it directly controls the balance between removing individual bias and preserving task-relevant features. We varied $\pi$ over non-uniform intervals [0.0, 0.1, 0.3, 0.6, 1.0] and visualize the results in Figure \ref{parameter}.

The sensitivity curves reveal distinct behaviors for each physiological indicator. Both HR and RR estimation benefit from a moderate orthogonality threshold. MAE decreases from $\pi=0.0$ to $\pi=0.3$, with the lowest errors at $\pi=0.3$, indicating that partial disentanglement of individual-specific information improves generalization. Beyond this point, increasing $\pi$ harms performance, as overly strong constraints suppress useful individual variation. In contrast, SpO2 estimation degrades steadily as $\pi$ increases, suggesting that oxygen saturation prediction relies more heavily on person-specific cues and is less tolerant of aggressive de-biasing. It is worth noting that, we fixed $\pi=0.1$ based on validation performance. While $\pi=0.1$ delivers consistently strong results across domains, it does not always achieve the absolute optimal MAE for every task (e.g., HR and RR achieve their best at $\pi=0.3$). We therefore acknowledge that a single global choice of $\pi$ represents a compromise: it offers stable, near-optimal performance without seeing the target domain, but target-domain-specific tuning of $\pi$ could yield further gains. 

\begin{figure*}
\begin{center}
\includegraphics[scale=0.11]{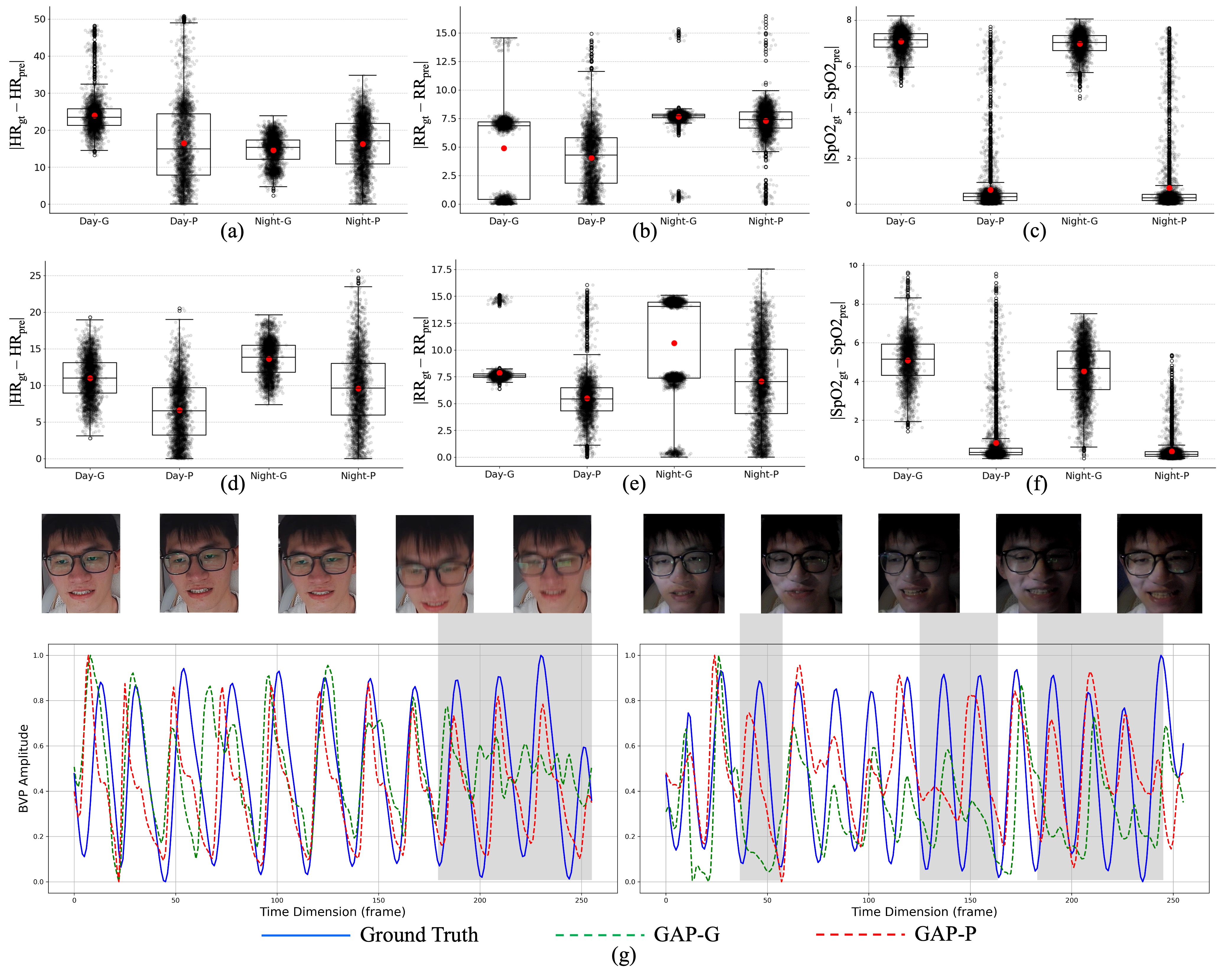}\\
\end{center}
\caption{The visualizations of our model's performance of all physiological estimation tasks on HMPC-D. The visualization results come from randomly selecting two individuals (one male and one female). Subfigures (a) to (c) belong to one subject and (d) to (g) belong to the other one. Subfigure (a) to (f) presents the distribution of the absolute error between the output of our model and the true value for the daytime and nighttime scenarios, respectively, in the form of box plot. The suffixes G and P stand for GAP-G and GAP-P that with personalized adaptation based on GAP-G, respectively. The red dots in each box represent the mean values and the black dots represent the specific deviation data. In subfigure (g), we provide the visualization of the ground-truth (blue line), output signal by GAP-G (green dashed line), and output signal by GAP-A (red dashed line) of one subject in daytime and night. We also highlight failure segments with grey.}\label{visual}
\end{figure*}

\begin{figure*}
\begin{center}
\includegraphics[scale=0.38]{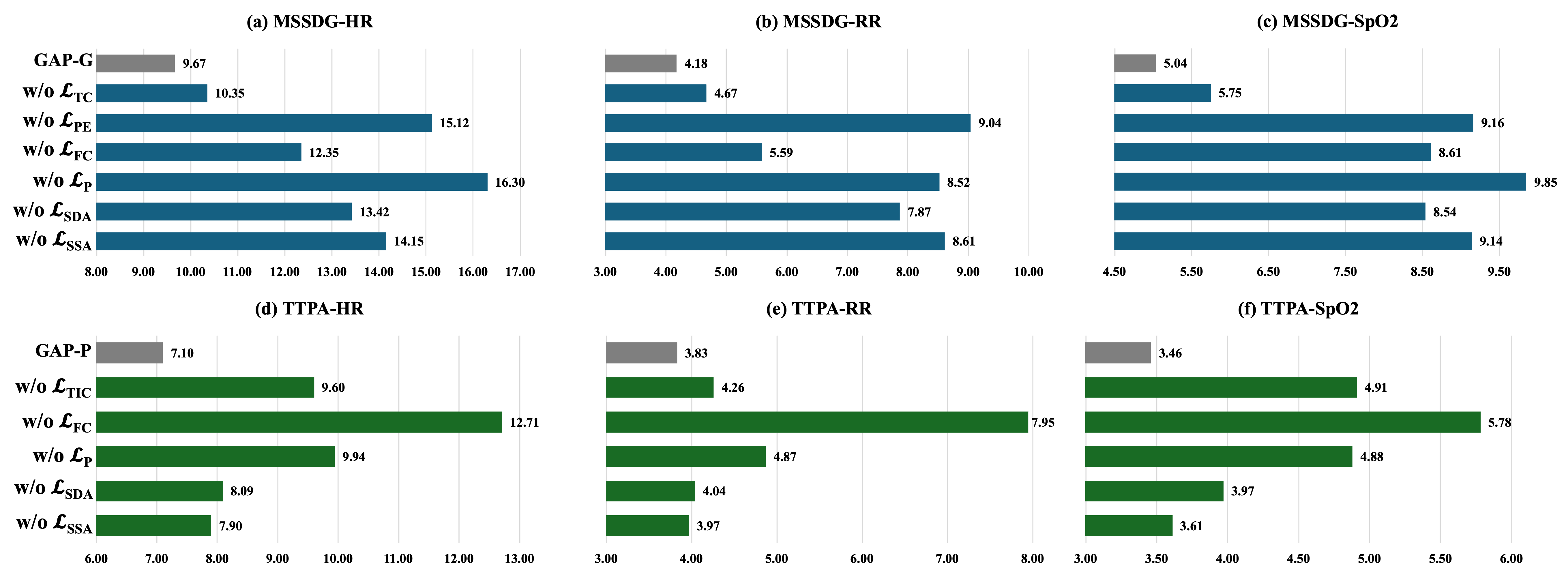}\\
\end{center}
\caption{MAE of applying different regularizations over GAP-G and GAP-P in the HMPC-D.}\label{ablation}
\vspace{-2mm}
\end{figure*}

\subsubsection{Evaluation of Personalization}
Based on the trained GAP-G, we independently adapted it to each individual, and conducted the evaluation at different illumination conditions. The results are presented in Table \ref{t6} and Figure \ref{visual}. First, GAP-P achieved the lowest RMSE values over all tasks during the day, showing significant improvement with personalization. For the HR estimation at night, GAP-P reduced the RMSE further to 7.27, illustrating enhanced accuracy through personalized adaptation. Overall, GAP-P maintained superior performance with lower RMSE and higher p, demonstrating the effectiveness of personalization in all scenarios.

Particularly, based on the absolute error distribution for the physiological estimation of two subjects (Figure \ref{visual}(a) to (f)), we notice that for both subjects, the GAP-P showed a reduction in the median and mean absolute errors compared to GAP-G. A similar trend was observed during nighttime scenarios, suggesting the robustness of the model in varying lighting conditions. Besides, as shown in Figure \ref{visual}(c)(f), the reduction in the interquartile range in the personalized adaptation setting indicated enhanced reliability and well-learned personal variability in predictions. Furthermore, we find GAP-G performed differently in these two subjects. After the personalization with GAP-P, the performance variation was reduced, particularly in RR and SpO2 (i.e., the error distribution is closer). Nevertheless, due to the effect of other uncontrollable environmental factors (e.g., head motions, dumpy road) in the real environment, we cannot determine that this performance change is brought about by individual heterogeneity. Therefore, the datasets collected in real driving environments in the future should attempt to control these environmental factors through experimental design for more comprehensive evaluations.  

Figure \ref{visual}(g) analyzes BVP signals during daytime and night. When comparing the blue line (ground truth) to GAP-G (green dashed) and GAP-P (red dashed), GAP-P showed better alignment with ground truth in most cases, no matter of peak number and amplitude, indicating superior tracking of both time and frequency-domain signals. A similar trend at night suggests that personalized adaptation enhanced error metrics and improved temporal fidelity in physiological signal estimation. Nevertheless, we have observed some failure cases in our analysis. In the selected daytime scene, the driver encountered two bumpy road sections, which resulted in continuous unclear human faces appearing in the video captured by the RGB camera. We note that GAP-G failed to restore the BVP signal in this scenario, while GAP-P maintained its robustness. We believe this discrepancy arises because the training source domain did not include such variations. Through the adaptation process, GAP-P effectively transfers the knowledge acquired in the source domain to the current distribution. In the night scene, we found that the light sources rely primarily on roadside illumination. This situation leads to challenges such as unbalanced facial brightness, unstable light sources, and head movements, which contribute to more failure cases for GAP-G. Although GAP-P has shown some improvement in this context, the limited and unstable lighting continues to pose difficulties. Specifically, the waveform output by GAP-P often struggles to match the ground truth in terms of signal morphology.

\subsubsection{Ablation test}
We also performed ablation experiments on the HMPC-D to validate the effectiveness of each of our module designs. As shown in Figure \ref{ablation} (a) to (c), we first found that independently eliminating the relevant modules for individual modeling ($\mathcal{L}_{PE}$ and $\mathcal{L}_{P}$) had a serious negative impact on model performance.  Eliminating $\mathcal{L}_{P}$ but performing $\mathcal{L}_{PE}$ caused the main task features to be orthogonal to the cluttered feature space, which could cause unnecessary interference in the main task feature space. In contrast, eliminating $\mathcal{L}_{PE}$ but executing $\mathcal{L}_{P}$ could reduce the generalization of the model by overly reinforcing the individual bias. The smallest impact was $\mathcal{L}_{TC}$, which aimed to improve time-domain consistency for generalization, but it also caused the model performance to degrade to the PhysMLE equivalent.

In the TTPA protocol (subfigures (d) to (f)), the most important constraint term was $\mathcal{L}_{FC}$, which was used as the most important source of gradients for all main tasks in the unsupervised condition. Reducing the $\mathcal{L}_{FC}$, the model would experience a severe degradation in performance under overemphasizing inconsistency and individual bias, compared to GAP-G. Besides, the necessity of individual bias ($\mathcal{L}_{P}$) as well as time-domain inconsistency $\mathcal{L}_{TIC}$ in the personalization process has also been demonstrated in the results.

\section{Limitation}
This work has several limitations that are worth noting. Firstly, compared to the latest work in MSSDG \citep{wang2024physmle}, the performance of our proposed method, GAP-G, has not shown significant improvement across all domains and tasks. We believe this may suggest that deep-based rPPG approaches based on large-scale supervised pre-training could have performance bottlenecks. These bottlenecks may arise not only from the low signal-to-noise ratio in the input data and inter-domain shifts but also from disparities on the label side. Specifically, variations can occur due to differences in the devices providing ground-truth signals across different datasets or the methods researchers use to calculate physiological indicators. For instance, the ground-truth HR provided in various datasets may be based on averages calculated over different time windows. Therefore, we recommend that future research should focus on addressing the discrepancies on the label side or explore combined approaches that integrate large-scale unsupervised training with small-scale, high-quality supervised fine-tuning.

Secondly, while the method we proposed has shown improved performance following generalizable training, our results on the real vehicle dataset remain limited. We believe this limitation is primarily due to dynamic natural lighting conditions in the driving environment and increased head movements. Additionally, the HMPC-D dataset we collected has a relatively small number of individuals and is affected by uncontrolled sun angles and weather conditions throughout the day, which may make it less representative for supervised training. We look forward to conducting a larger and better-designed data collection experiment in the future.

\section{Conclusion}\label{sec13}

In this paper, we presented a unified framework, GAP, for multi-task remote physiological measurement that effectively addresses the challenges of MSSDG and TTPA. Our approach leveraged insights from biometrics and rPPG methodologies to enhance the estimation of multiple vital signs (i.e., BVP, HR, RR, and SpO2). From a structural perspective, our proposed framework introduced novel input-level data augmentation techniques and self-supervised regularization methods that incorporated prior knowledge from RoR-based SpO2 estimation, semantic structure and distribution alignment. It not only further mitigated the shortage of our preliminary work \citep{wang2024physmle}, but also improved the robustness of our model against semantic-irrelevant noise. Furthermore, we emphasized the importance of individual heterogeneity in physiological estimation by introducing a flexible modeling scheme that consisted of explicit individual bias modeling and spatio-temporal time-domain inconsistency. It
allowed for the adjustment of individual biases, and facilitated personalized adaptations in MTL settings. Our extensive experimental validation across six public datasets, and one newly collected HMPC-D dataset for monitoring in driving, demonstrated the superiority of the proposed GAP in both MSSDG and TTPA protocols. The results indicated substantial improvements in physiological signal estimation accuracy and computational cost when compared to existing methods, underscoring the effectiveness of our proposed framework in real-world scenarios. 

Given the different findings about the effect of illumination derived from the test on HMPC-D, future research should focus on refining MTL monitoring in more challenging real-world scenarios (e.g., driving). Additionally, the principles of MSSDG and TTPA introduced in this work can be applied to other multi-task domains, such as salient object detection \citep{sun2024defense} and semantic segmentation \citep{wang2022semi}. Moreover, the idea of leveraging a unified framework to achieve both generalization and personalization based on small adjustments also holds promise for future studies and real-world practice. 

\backmatter

\bmhead{Acknowledgements}

This work was supported by the Natural Science Foundation of Guangdong Province of China (2024A1515010392), and partially by the Technology Project (No. 2023A03J0011) and Guangdong Provincial Key Lab of Integrated Communication, and Sensing and Computation for Ubiquitous Internet of Things (No. 2023B1212010007).

\section*{Declarations}
There is a comprehensive portal to find these five datasets (i.e., UBFC, BUAA, PURE, VIPL-HR, V4V) source sites https://github.com/EnVision-Research/NEST-rPPG. For HCW and HMPC-D, the authors declare that the data supporting the findings of this study are available from the corresponding author upon reasonable request.

\bibliography{bibliography}

\end{document}